\documentclass[10pt]{article} 
\usepackage{times}
\usepackage[left=1.5in, right=1.5in, top=1in, bottom=1in]{geometry}
\usepackage{breakcites}
\usepackage{graphicx}
\usepackage{textcomp}
\usepackage{xcolor}
\def\BibTeX{{\rm B\kern-.05em{\sc i\kern-.025em b}\kern-.08em
    T\kern-.1667em\lower.7ex\hbox{E}\kern-.125emX}}

\usepackage{amsmath}
\usepackage[]{algorithm2e}
\usepackage{algpseudocode}
\SetKwRepeat{Do}{do}{while}
\usepackage{graphicx}  
\usepackage{float}
\usepackage{bm}      
\usepackage{epstopdf}
\usepackage{subfig}
\usepackage{amssymb}   
\usepackage{bm}
\usepackage{booktabs}       
\usepackage{hyperref}

\DeclareMathOperator{\diag}{diag}

\begin{document}

\title{Towards Deterministic Diverse Subset Sampling}

\author{Joachim Schreurs, Micha\"el Fanuel and  Johan A. K. Suykens\\
Department of Electrical Engineering, ESAT-STADIUS,\\
KU Leuven. Kasteelpark Arenberg 10, B-3001 Leuven, Belgium\\
\texttt{\{joachim.schreurs,michael.fanuel,johan.suykens\}@kuleuven.be} \\
}

\maketitle          
\begin{abstract}
Determinantal point processes (DPPs) are well known models for diverse subset selection problems, including recommendation tasks, document summarization and image search. In this paper, we discuss a greedy deterministic adaptation of k-DPP. Deterministic algorithms are interesting for many applications, as they provide interpretability to the user by having no failure probability and always returning the same results. First, the ability of the method to yield low-rank approximations of kernel matrices is evaluated by comparing the accuracy of the Nystr\"om approximation on multiple datasets. Afterwards, we demonstrate the usefulness of the model on an image search task.
\end{abstract}
\section{Introduction}

Selecting a diverse subset is an interesting problem for many applications. Examples are document or video summarization~\cite{carbonell1998use,gong2014diverse,kulesza2010structured,kulesza2011learning}, image search tasks~\cite{kulesza2011k}, pose estimation~\cite{kulesza2010structured} and many others. Diverse sampling algorithms have also shown their benefits to calculate a low-rank matrix approximations using the Nystr\"om method~\cite{williams2001using}. This method is a popular tool for scaling up kernel methods, where the quality of the approximation relies on selecting a representative subset of landmark points or Nystr\"om centers. 

\paragraph{Notations}
In this work, we will use uppercase letters for matrices and calligraphic letters for sets, while bold letter denote random variables. The notation  $(\cdot)^\dagger$ denotes Moore-Penrose pseudo inverse of a matrix. We also define the partial order of positive definite (resp. semidefinite) matrices by $A\succ B$ (resp. $A\succeq B)$ if and only if $A-B$ is positive definite (resp. semidefinite). Furthermore, we denote by $K$, the Gram matrix $[k(x_i,x_j)]_{i,j = 1}
^n$ obtained from a positive semidefinite kernel such as the Gaussian kernel $k(x,y) = \exp(-\|x-y\|_2^2/(2\sigma^2))$.
\paragraph{Nystr\"om approximation}
The Nystr\"om method takes a positive semidefinite matrix $K \in \mathbb{R}^{n\times n}$ as input, selects from it a small subset $\mathcal{C}$ of columns, and constructs the approximation $\hat{K}=K_{ \mathcal{C}} K_{\mathcal{C} \mathcal{C}}^{\dagger} K_{\mathcal{C} }^{\top}$, where  $K_{\mathcal{C}}=K C$ and $K_{\mathcal{C} \mathcal{C} }=C^{\top} K C$ are submatrices of the kernel matrix and $C \in \mathbb{R}^{n \times|\mathcal{C}|}$ is a sampling matrix obtained by selecting the columns of the identity matrix indexed by $\mathcal{C}$. 
The matrix $\hat{K}$ is used in the place of $K$, so to decrease the training runtime and memory requirements. Using a dependent or \emph{diverse} sampling algorithm for the  Nystr\"om approximation has shown to give better performance than independent sampling methods in~\cite{RegularizedChristoffel,Fastdpp}. 
\paragraph{Determinantal Point Processes and kernel methods}
Determinantal point processes (DPPs)~\cite{KuleszaT12} are well known models for diverse subset selection problems. A point process  on a ground set $[n] = \{1,2,...,n\}$ is a probability measure over point patterns, which are finite subsets of $[n]$. 
It is common to define a DPP thanks to its marginal kernel, that is a positive symmetric semidefinite matrix satisfying $P \preceq \mathbb{I}$.
Let $\bm{\mathcal{Y}}$ denote a random subset, drawn according to the DPP with marginal kernel $P$. Then, the probability that $\mathcal{C}$ is a subset of the random $\bm{\mathcal{Y}}$ is defined by
\begin{equation}
\label{eq:origDPP}
\mathrm{Pr}(\mathcal{C} \subseteq \bm{\mathcal{Y}}) = \mathrm{det}(P_{\mathcal{C}\mathcal{C}}).
\end{equation}
 Notice that all principal submatrices of a positive semidefinite matrix are positive semidefinite.  From \eqref{eq:origDPP}, it follows that:
\begin{align*}
\mathrm{Pr}(i \in \bm{\mathcal{Y}}) &= P_{ii} \\
\mathrm{Pr}(i,j \in \bm{\mathcal{Y}}) &= P_{ii}P_{jj} - P_{ij}P_{ji}\\
\label{eq:KDPP}
&= \mathrm{Pr}(i \in \bm{\mathcal{Y}})\mathrm{Pr}(j \in \bm{\mathcal{Y}}) - P_{ij}^2.
\end{align*}
The diagonal elements of the kernel matrix give the marginal probability of inclusion for individual elements, whereas the off-diagonal elements determine the "repulsion" between pairs of elements. Thus, for large values of $P_{ij}$, or a high similarity, points are unlikely to appear together. In some applications, it can be more convenient to define DPPs thanks to $L$-ensembles, which can be related to marginal kernels by the formula $L = P(\mathbb{I} - P)^{-1}$ when $P\prec \mathbb{I}$ (more details in~\cite{borodin2009determinantal}). They allow to define the probability of sampling a random subset $\bm{\mathcal{Y}}$ that is equal to $\mathcal{C}$:
\begin{equation}
\mathrm{Pr}(\bm{\mathcal{Y}} = \mathcal{C}) = \frac{\mathrm{det}(L_{\mathcal{C}\mathcal{C}})}{\mathrm{det}(\mathbb{I} + L)}.\label{eq:Lensemble}
\end{equation}
 In contrast to \eqref{eq:origDPP}, the only requirement on $L$ is that it has to be positive semidefinite. Notice that the normalization in \eqref{eq:Lensemble} can be derived classically by considering the property relating the coefficients of the characteristic polynomial of a matrix to the sum of the determinant of its  principal submatrices of the same size. In this paper, the $L$-ensemble is chosen to be a kernel matrix $K$.

Exact sampling of a DPP is done in two phases~\cite{KuleszaT12}. Let $V$ be the matrix whose columns are the eigenvectors of $K$. In the first phase, a subset of eigenvectors of the kernel matrix $K$ is selected at random, where the probability of selecting each eigenvector depends on its associated eigenvalue in a specific way given in Algorithm \ref{AlgDPP}. In the second phase, a sample $\mathcal{Y}$ is produced based on the selected vectors. At each iteration of the second loop, the cardinality of $\mathcal{Y}$ increases by one and the number of columns of $V$ is reduced by one. A k-DPP~\cite{kulesza2011k} is  a DPP conditioned on a fixed cardinality $|\mathcal{Y}| = k$. Note that $e_i\in \mathbb{R}
^n$ is the $i$-th vector of the canonical basis. 	

\begin{algorithm}[h]
	\begin{algorithmic}[1]
		\Statex {\bf input}: $L$-ensemble $L\succeq 0$
		\Statex {\bf initialization}: $\mathcal{J} = \emptyset$ and $\mathcal{Y} = \emptyset$ 
		\Statex Calculate the eigenvector/value pairs $\{(v_i,\lambda_i)\}_{i=1}^n$ of $L$.
		\Statex {\bf for}: $i=1, \ldots, n$  {\bf do}
		\Statex \qquad $\mathcal{J} \leftarrow \mathcal{J} \cup\{i\} $ \text { with prob. }  $\frac{\lambda_{i}}{\lambda_{i}+1}$
		\Statex {\bf end for}
		\Statex $V \leftarrow\left\{v_{i}\right\}_{i \in \mathcal{J}}$ a set of columns
		\Statex {\bf while}: $|V| > 0$ \, {\bf do}
		\Statex \qquad Draw an index $i$ according to the distribution $p_i = \frac{1}{|V|} \sum_{v \in V} (v^\mathrm{T} e_i)^2.$
		\Statex \qquad $\mathcal{Y} \leftarrow \mathcal{Y} \cup i$
		\Statex \qquad $V \leftarrow V_\perp$, an orthonormal basis for the subspace of $V$ orthogonal to $e_i$.
		\Statex {\bf end while}
		\Statex {\bf return} $\mathcal{Y}$.
	\end{algorithmic} 
	\caption{Exact DPP sampling algorithm associated to the $L$-ensemble $L$~\cite{KuleszaT12}. Notice that $P = L(L+\mathbb{I})^{-1}$, so that the eigenvector/value pairs of $P$ are exactly $\{(v_i,\frac{\lambda_{i}}{\lambda_{i}+1})\}_{i=1}^n$. \label{AlgDPP}}
\end{algorithm}

Deterministic algorithms are interesting for many applications, as they provide interpretability to the user by having no chance of failure and always returning the same results. The usefulness of deterministic algorithms has already been recognized by Papailiopoulos et al.~\cite{Papailiopoulos} and McCurdy~\cite{McCurdy}, who provide deterministic algorithms based on the (ridge) leverage scores. These statistical leverage scores  correspond to correlations between the singular vectors of a matrix and the canonical basis~\cite{alaoui2015fast,drineas2012fast}. The recently introduced Deterministic Adaptive Sampling (DAS) algorithm~\cite{RegularizedChristoffel} provides a deterministically obtained \emph{diverse} subset. The method shows superior performance compared to randomized counterparts in terms of approximation error for the Nystr\"om approximation when the eigenvalues of the kernel matrix have a fast decay. A similar observation was made for the deterministic algorithms of Papailiopoulos et al.~\cite{Papailiopoulos} and McCurdy~\cite{McCurdy}. \\  

This paper discusses a deterministic adaption of k-DPP, where we have the following empirical observations:
\begin{enumerate}
	\item The method is deterministic, hence there is no failure probability and the method always produces the same output. 
	\item Only the $k$ eigenvectors with the largest eigenvalues are needed, which results in a speedup when $k \ll n$.
	\item We observed that the method samples a more diverse subset than the original k-DPP on multiple datasets.
	\item There is no need to tune a regularization parameter, which is the case for the DAS algorithm.
	\item The method shows superior accuracy in terms of the max norm of the Nystr\"om approximation on multiple datasets compared to the standard k-DPP, along with better accuracy of the kernel approximation for the operator norm when there is fast decay of the eigenvalues.
\end{enumerate}

In Section \ref{sec:DetkDPP}, we introduce the method. Secondly, we make a connection with the DAS algorithm, namely the  deterministic k-DPP corresponds to the DAS algorithm with an adapted projector kernel matrix. In Section \ref{sec:numRes}, we evaluate the method on different datasets. Finally, a small real-life illustration is shown in Section \ref{sec:application}.

\section{Deterministic adaptation of k-DPP \label{sec:DetkDPP}}

We discuss a deterministic adaptation of k-DPP, by selecting iteratively landmarks with the highest probability. As it is described in Algorithm \ref{AlgProposed}, we can successively maximize the probability over a nested sequence of sets $\mathcal{C}_{0} \subseteq \mathcal{C}_{1} \subseteq \cdots \subseteq \mathcal{C}_{k}$ starting with $\mathcal{C}_{0}=\emptyset$ by adding one landmark at each iteration. The proposed method is an adaptation of the improved k-DPP sampling algorithm given by Tremblay et al.~\cite{tremblay2018optimized}. Our proposed method start from a projective marginal kernel $P = VV^\top$, with $V = [v_1,...,v_k] \in \mathbb{R}^{n \times k}$ the sampled eigenvectors of the kernel matrix. Instead of sampling the eigenvectors~\cite{KuleszaT12}, the $k$ eigenvectors with the largest eigenvalue are chosen. Secondly, at each iteration the point with the highest probability $p(i)= P_{ii}-P_{\mathcal{C} i}^{\top} P_{\mathcal{C}\mathcal{C}}^{\dagger} P_{\mathcal{C} i}$ is chosen, where $\mathcal{C}$ corresponds to the selected subset so far. Besides the interpretation of DPPs in relation to diversity, the aforementioned probability gives a second insight in diversity. Namely we have $p(i)= \left\|v_{i}-\pi_{V_{\mathcal{C}}} v_{i}\right\|_{2}^{2},$ where $v_{i} \in \mathbb{R}^{n \times 1}$ is the $i$-th column of $V$ and $\pi_{V_ {\mathcal{C}}}$ is the projector on $V_{\mathcal{C}}=\operatorname{span}\left\{v_{s} | s \in \mathcal{C}\right\}$. The chosen landmark corresponds to the point that is the most distant to the space of the previously sampled points.

\begin{algorithm}[h]
	\begin{algorithmic}[2]
		\Statex {\bf input}: Kernel matrix $K$, sample size $k$.
		\Statex {\bf initialization}:  $\mathcal{C} \leftarrow \emptyset$
		\Statex Calculate the first $k$ eigenvectors $\mathrm{V} \in \mathbb{R}^{n \times k}$  from $K$.
		\Statex $P = VV^\mathrm{T}$
		\Statex Define $p_{0} \in \mathbb{R}^{N} : \forall i, \quad p_{0}(i)=\left\|\mathrm{V}^{\top} e_{i}\right\|^{2}$
		\Statex $p \leftarrow p_{0}$
		\Statex {\bf for}: $i=1, \ldots, k$  {\bf do}
		\Statex \qquad Select $c_{i}$ with highest probability $p(i)$
		\Statex \qquad $\mathcal{C} \leftarrow \mathcal{C} \cup\left\{c_{i}\right\}$
		\Statex \qquad Update $\boldsymbol{p} : \forall j \quad p(j)=p_{0}(j)-P_{\mathcal{C} j}^{\top} P_{\mathcal{C}\mathcal{C}}^{\dagger} P_{\mathcal{C} j}$
		\Statex {\bf end for}
		\Statex {\bf return} $\mathcal{C}$.
	\end{algorithmic} 
	\caption{Deterministic adaptation of the k-DPP sampling algorithm.\label{AlgProposed}}
\end{algorithm}

\subsection{Connections with DAS}

Algorithmically, the proposed method corresponds to the DAS algorithm~\cite{RegularizedChristoffel} (see, Algorithm \ref{DAS} in appendix) with a different projector kernel matrix.  More precisely, DAS uses a smoothed projector kernel matrix $P_{n \gamma}(K)=K(K+n \gamma I)^{-1}$ with $K\succ 0$. The ridge leverage scores $l_{i}(\gamma)=\sum_{j=1}^{n} \frac{\lambda_{j}}{\lambda_{j}+n \gamma} V_{i j}^{2}$ can be found on the diagonal $P_{n \gamma}(K)$. Let $V \in \mathbb{R}^{n \times n} $ be the matrix of eigenvectors of the kernel matrix $K$. On the contrary, in this paper, the proposed method has a \emph{sharp projector kernel matrix} with the rank-$k$ leverage scores $l_{i}= \sum_{j=1}^{k} V_{i j}^{2}$ on the diagonal. This has the added benefit that there is no regularization parameter to tune. The DAS algorithm is given in the Appendix. 

\begin{figure}[h]
 	\centering
 	\subfloat[Uniform]{\label{fig:ExampleUnif}\includegraphics[width=0.3\textwidth]{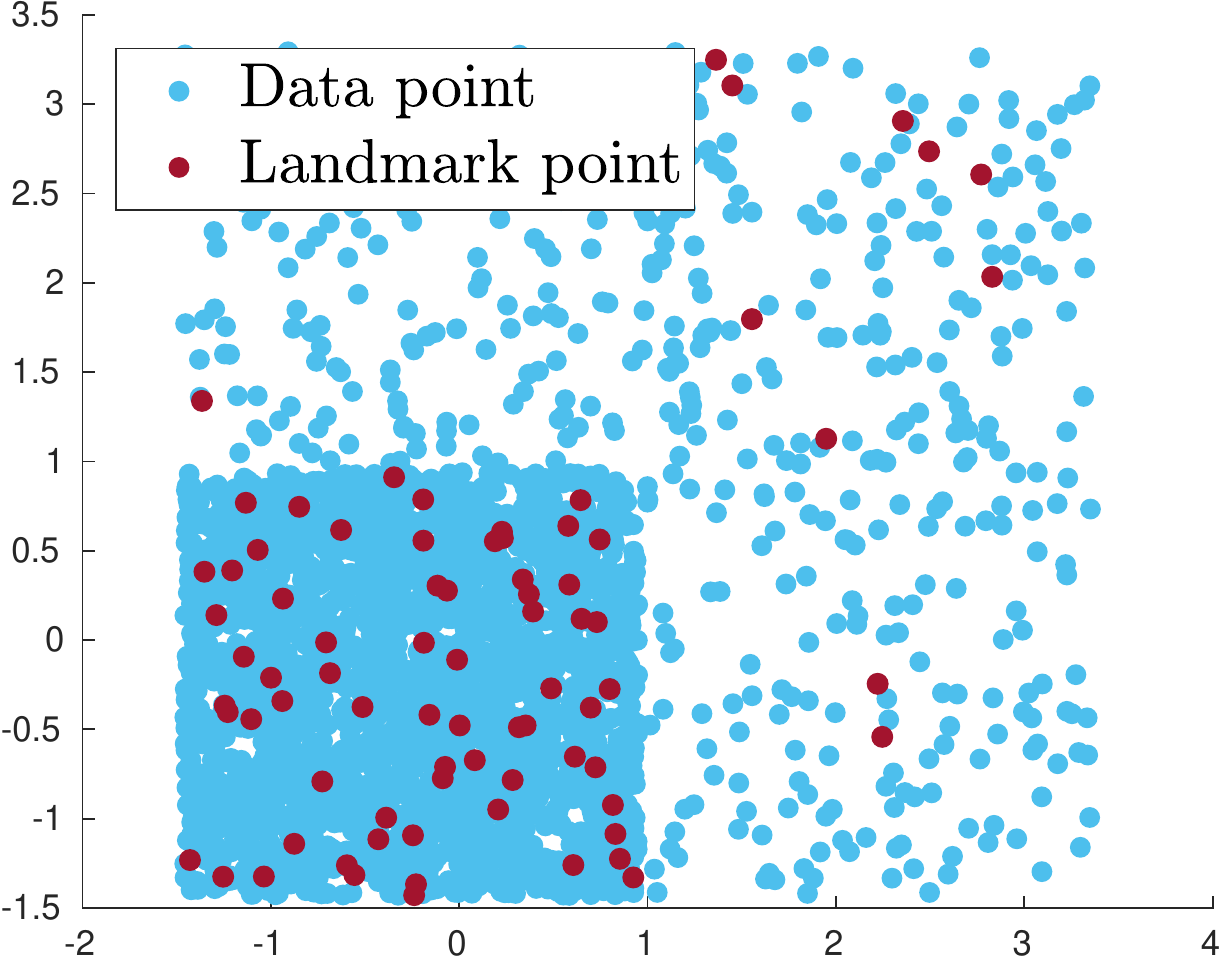}}   
 	\quad
 	\subfloat[k-DPP]{\label{fig:ExamplekDPP}\includegraphics[width=0.3\textwidth]{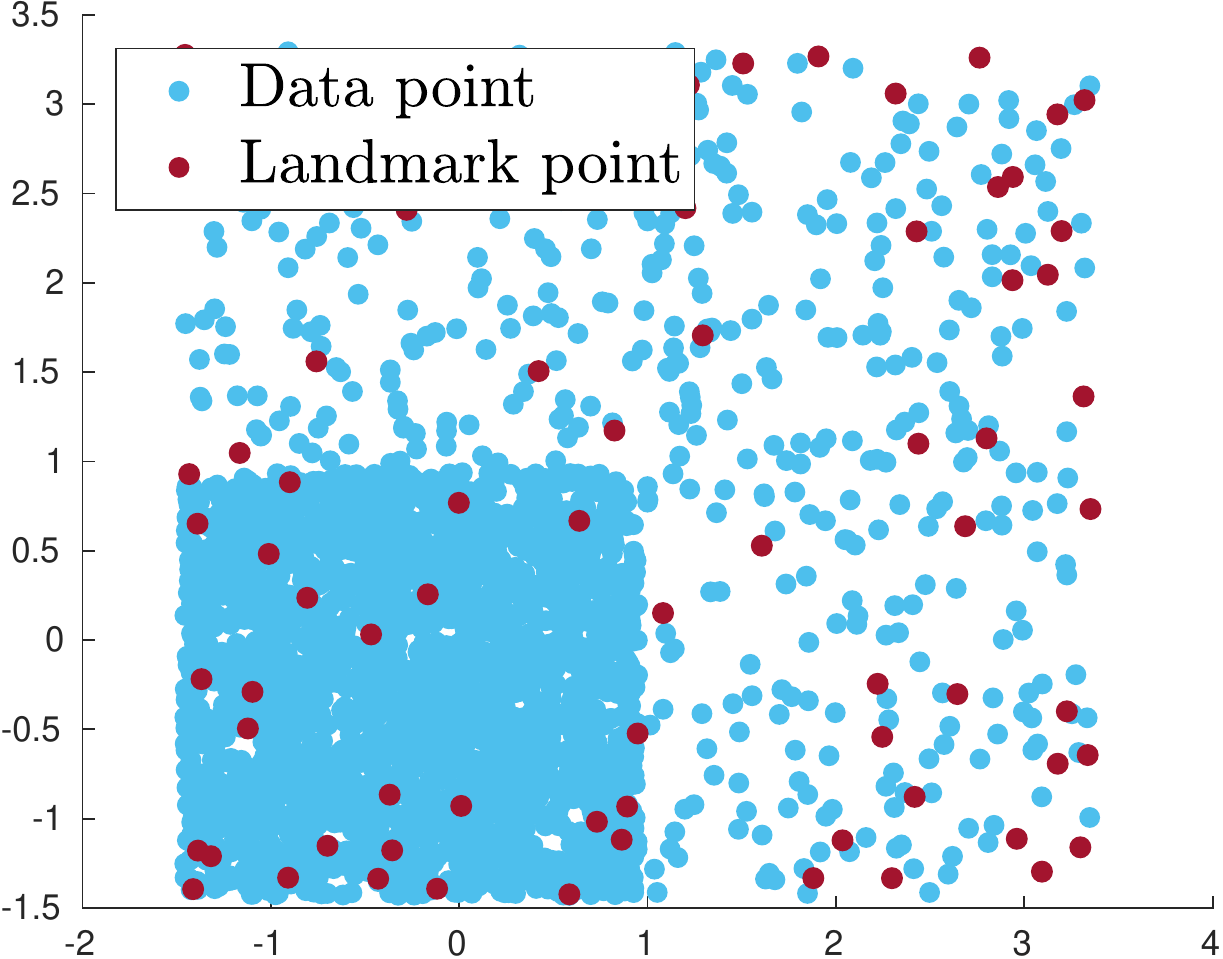}}   
 	\quad
 	\subfloat[Deterministic k-DPP]{\label{fig:ExampleDerkDPP}\includegraphics[width=0.3\textwidth]{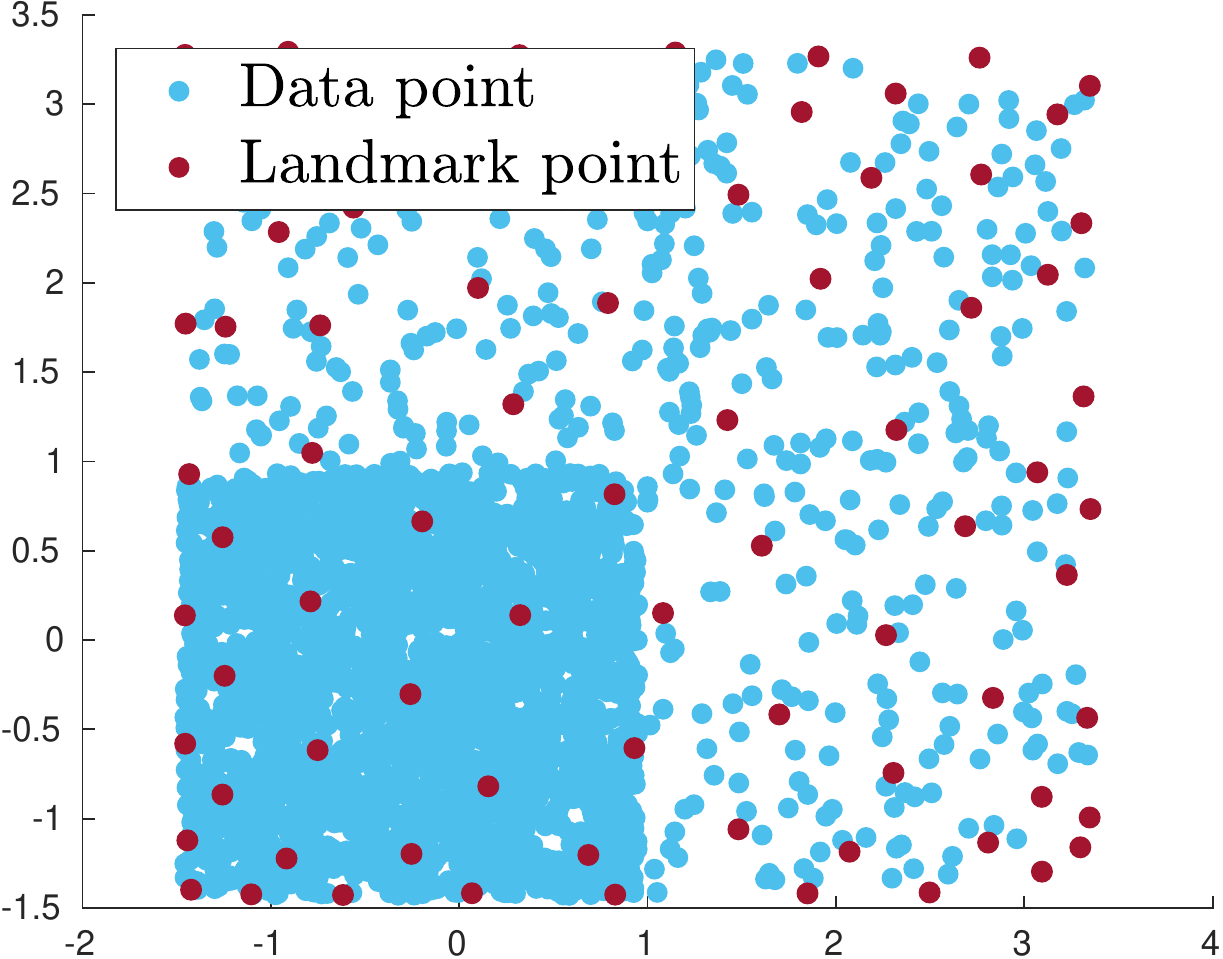}}   
 	\caption{Illustration of sampling methods on an artificial dataset. Uniform sampling does not promote diversity and selects almost all points in the bulk of the data. Sampling a k-DPP overcomes this limitation, however landmarks can be close to each other. The latter is solved by using the deterministic adaptation of k-DPP, which gives a more diverse subset.}\label{fig:Example}
 \end{figure}

Algorithm \ref{AlgProposed} is a greedy reduced basis method as defined in \cite{devore2013greedy}. Other greedy maximum volume approaches are found in \cite{chen2018fast,ccivril2009selecting,gillenwater2012near}. These greedy methods are used for finding an estimate of the most likely configuration (MAP), which is known to be NP-hard~\cite{gillenwater2012near}. In practice, we see that the method performs quite well and gives a consistently larger $\mathrm{det}(K_{\mathcal{C}\mathcal{C}})$, which is considered a measure for diversity, compared to the randomized counterpart (see Section \ref{sec:numRes}). This number is calculated as follows $\mathrm{log}(\mathrm{det}(K_{\mathcal{C}\mathcal{C}})) = \sum_{i=1}^k \mathrm{log}(\lambda_i)$, with $\{\lambda_i\}_{i=1}^k$ the singular values of $K_{\mathcal{C}\mathcal{C}}$.  A small illustration is given in Figure \ref{fig:Example}, where the deterministic algorithm gives a more diverse subset.

\section{Numerical results \label{sec:numRes}}

We evaluate the performance of the deterministic variant of the k-DPP with a Gaussian kernel on the \texttt{Boston housing}, \texttt{Stock},  \texttt{Abalone} and  \texttt{Bank 8FM} datasets, which have 506, 950, 4177 and 8192 datapoints respectively. Those public datasets\footnote{\url{https://www.cs.toronto.edu/~delve/data/datasets.html}, \url{https://www.openml.org/d/223}} have been used for benchmarking  k-DPPs in~\cite{Fastdpp}. The implementation of the algorithms is done with MatlabR2018b.

\begin{figure}[h]
	\centering
	\subfloat[$\mathrm{log(\mathrm{det}(K_{\mathcal{C}\mathcal{C}}))}$]{\label{fig:stock1_Det}\includegraphics[width=0.45\textwidth]{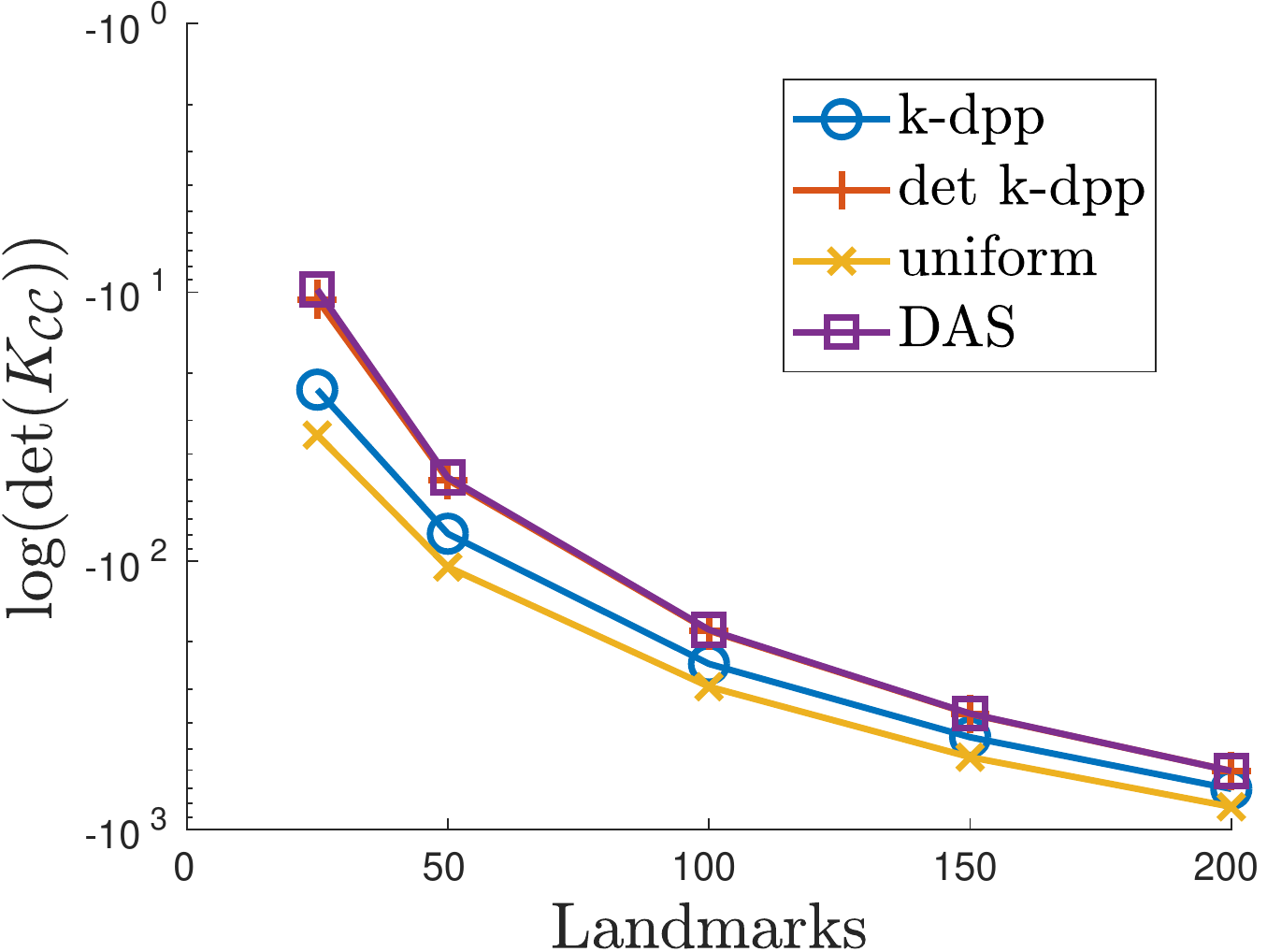}}   
	\quad
	\subfloat[operator norm]{\label{fig:Stock1_OP}\includegraphics[width=0.45\textwidth]{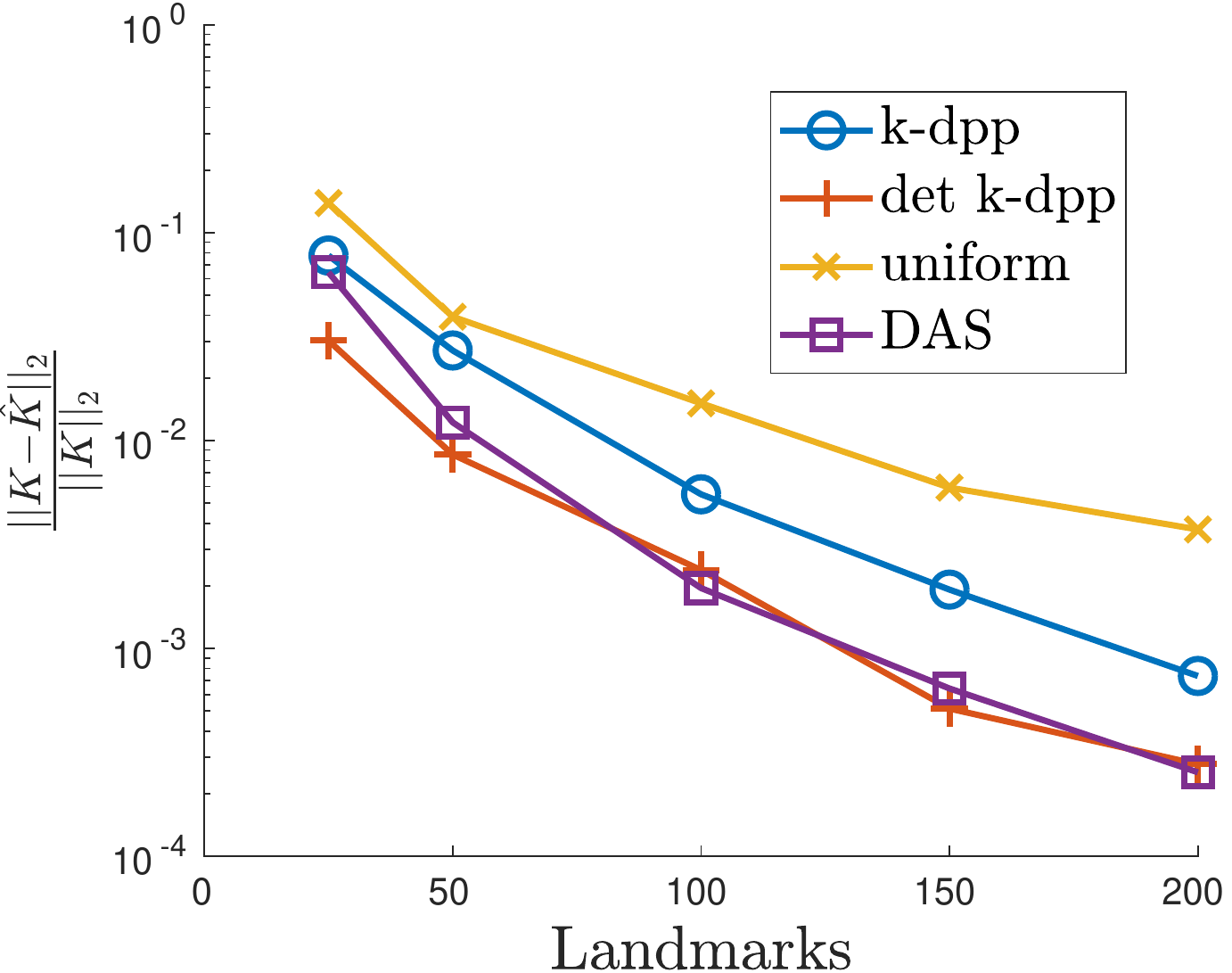}}   
	\quad
	\subfloat[max norm]{\label{fig:stock1_Max}\includegraphics[width=0.45\textwidth]{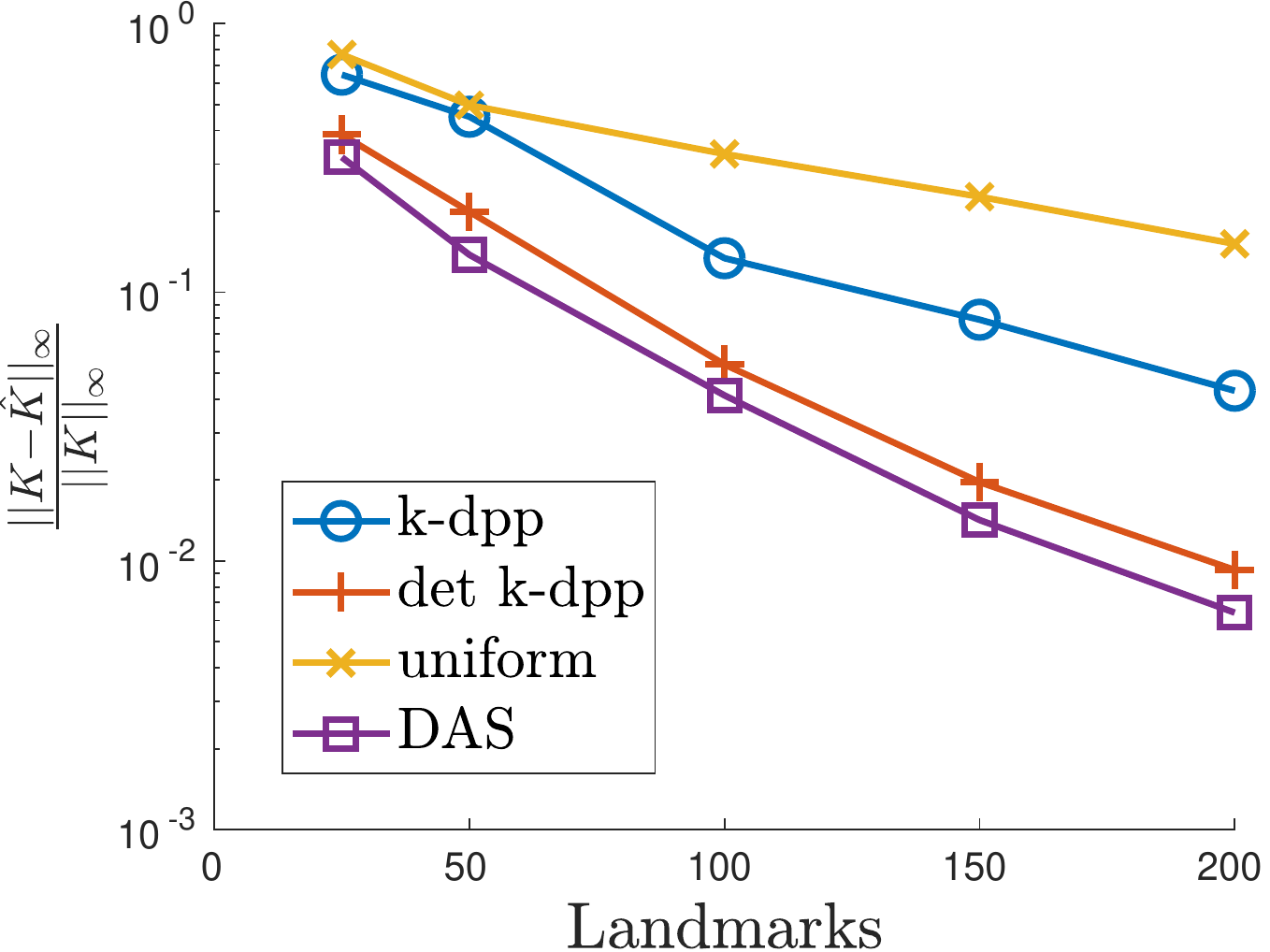}}   
	\quad
	\subfloat[timing]{\label{fig:stock1_Time}\includegraphics[width=0.45\textwidth]{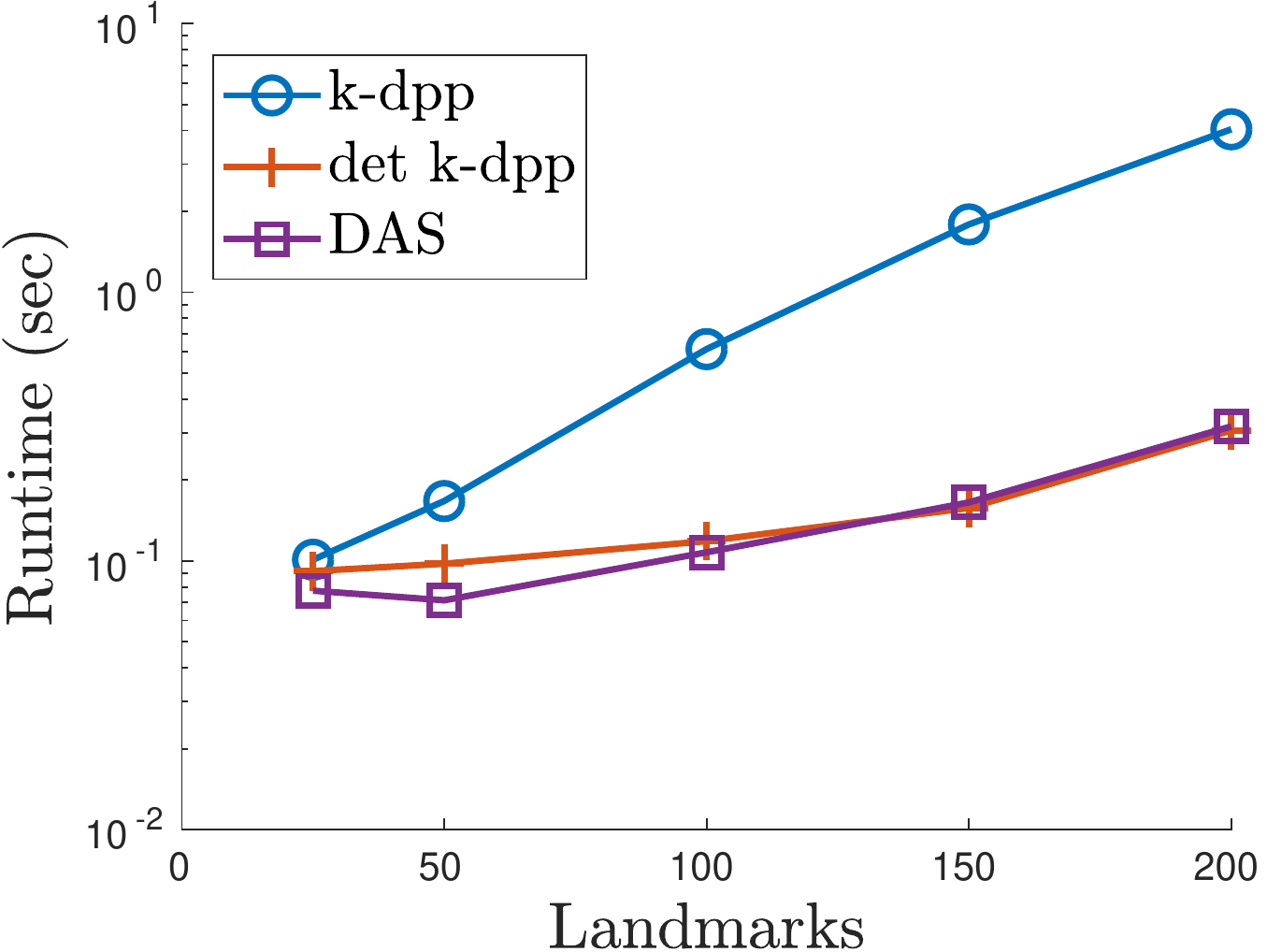}}   
	\caption{The $\mathrm{log}(\mathrm{det}(K_{\mathcal{C}\mathcal{C}}))$, relative operator norm and relative max norm of the Nystr\"om approximation error and timings as a function of the number of landmarks on the Stock dataset. The results are plotted on a logarithmic scale, averaged over 10 trials. The larger $\mathrm{log}(\mathrm{det}(K_{\mathcal{C}\mathcal{C}}))$, the more diverse the subset.}\label{fig:Stock}
\end{figure}

Throughout the experiments, we use a fixed bandwidth $\sigma = 2$ for \texttt{Boston housing}, \texttt{Stock} and \texttt{Abalone} dataset and $\sigma = 5$ for the \texttt{Bank 8FM} dataset after standardizing the data.  The following algorithms are used to sample $k$ landmarks: Uniform sampling, k-DPP\footnote{We used the Matlab code available at \url{https://www.alexkulesza.com/}.}~\cite{kulesza2011k}, DAS~\cite{RegularizedChristoffel} and the proposed method. DAS is executed for multiple regularization parameters $\gamma \in \{10^0 , 10^{-1} , \dots , 10^{-6} \}$ where the sample with the best performing $\gamma$ is selected to approximate the kernel matrix. The total experiment is repeated 10 times. 
The quality of the landmarks $\mathcal{C}$ is evaluated by the relative operator or max norm $\|K-\hat{K}\|_{\{\infty,2\}}/\|K\|_{\{\infty,2\}}$ with $\hat{K} = K_{ \mathcal{C}} (K_{\mathcal{C} \mathcal{C}} + \epsilon\mathbb{I})^{-1} K_{\mathcal{C}}^\top$ with $\varepsilon = 10^{-12}$ for numerical stability. The max norm and the operator norm of a matrix $A$ are given respectively by $\|A\|_{\infty}=\max _{i, j}\left|A_{i j}\right|$ and $\|A\|_{2}=\max _{\|x\|_{2}=1}\|A x\|_{2}$. The diversity is measured by $\mathrm{log}(\mathrm{det}(K_{\mathcal{C}\mathcal{C}}))$, where a larger log determinant means more diversity. The results for the \texttt{Stock} dataset are visible in Figure \ref{fig:Stock}.  The results for the rest of the datasets or shown in Figures~\ref{fig:Det}, \ref{fig:OP}, \ref{fig:Max} and \ref{fig:Time} in the Appendix. The computer used for these simulations has $8$ processors $3.40GHz$ and $15.5$ GB of RAM.

As previously mentioned, the greedy method returns a more diverse subset. Figure~\ref{fig:Det} shows the $\mathrm{log}(\mathrm{det}(K_{\mathcal{C}\mathcal{C}}))$ where the proposed method shows similar performance as DAS, while improving on both the randomized methods. The same is visible for the relative max norm of the Nystr\"om approximation error. DAS and the deterministic variant of the k-DPP perform well on the \texttt{Boston housing} and \texttt{Stock} dataset, which show a fast decay in the spectrum of $K$ (see Figure~\ref{fig:Eig}). If the decay of the eigenvalues is not fast enough, the randomized k-DPP, shows better performance. The same observation was made for the deterministic (ridge) leverage score sampling algorithms~\cite{McCurdy,Papailiopoulos} as well as DAS~\cite{RegularizedChristoffel}.

\section{Illustration \label{sec:application}}

We demonstrate the use of the proposed method on a image summarization task.
The first experiment is done on the \texttt{Stanford Dogs} dataset\footnote{\url{http://vision.stanford.edu/aditya86/ImageNetDogs/main.html}}, which contains images of 120 breeds of dogs from around the world. This dataset has been built using images and annotation from ImageNet for the task of fine-grained image categorization. The training features are SIFT descriptors~\cite{lowe1999object} (given by the dataset), which are used to make a histogram intersection kernel. We take a subset of 50 images each of the classes \textit{border collie}, \textit{chihuahua} and \textit{golden retriever}. The total training set is visualized in Figure \ref{fig:TrainData} in the Appendix. Figure \ref{fig:Dogs} displays the results of the method for $k=4$. One can observe that the images are very dissimilar and dogs out of each breed are represented. This is confirmed by the projection of the landmarks on the 2 first principal components of the kernel principal component analysis (KPCA)~\cite{scholkopf1997kernel}, where the landmark points lie in the outer regions of the space.

 \begin{figure}[h]
 	\centering
 	\begin{minipage}{.45\linewidth}
 	\subfloat[]{\label{fig:Sample1}\includegraphics[width=0.42\textwidth]{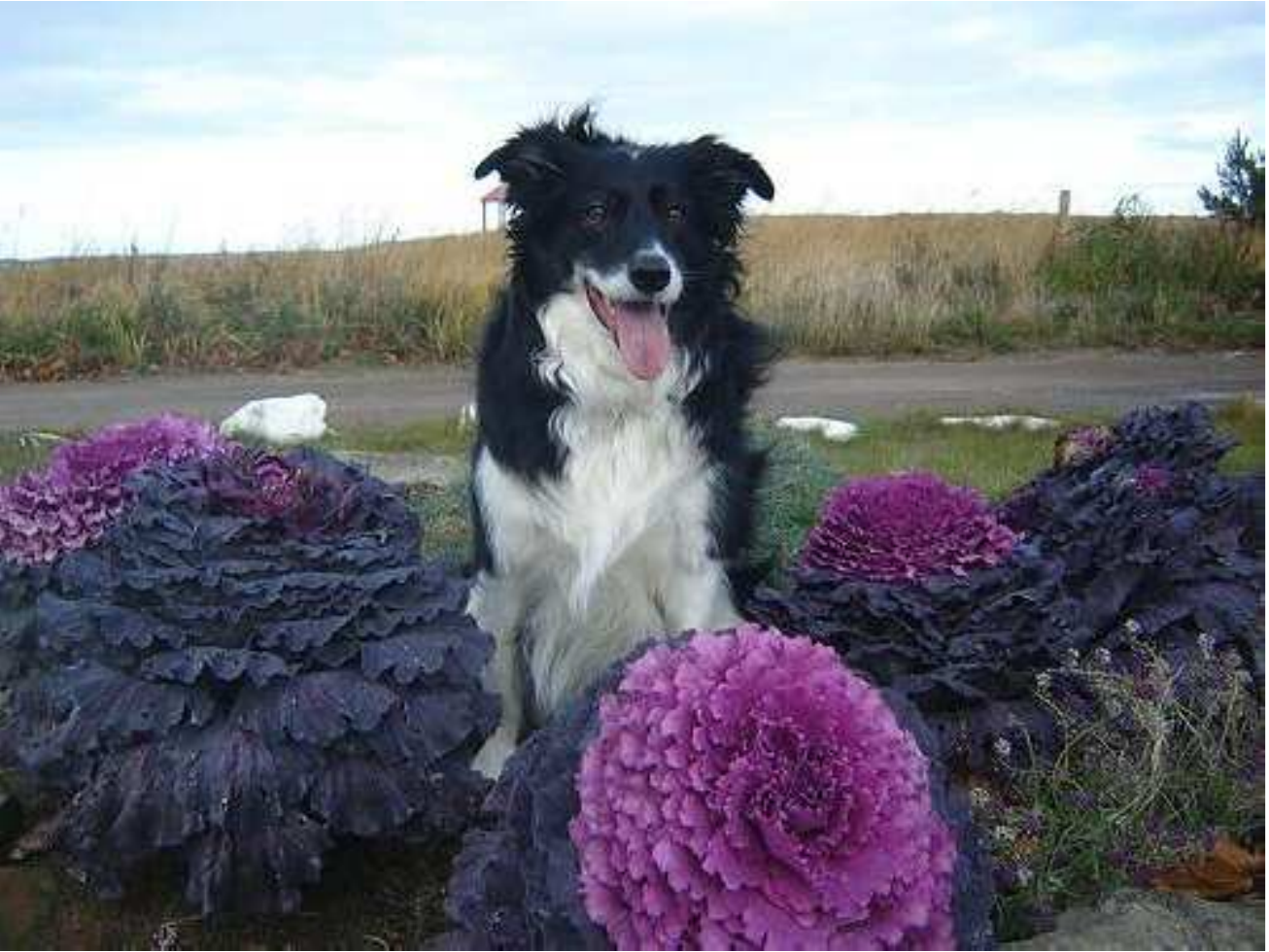}}
 	\quad  
 	\subfloat[]{\label{fig:Sample2}\includegraphics[width=0.42\textwidth]{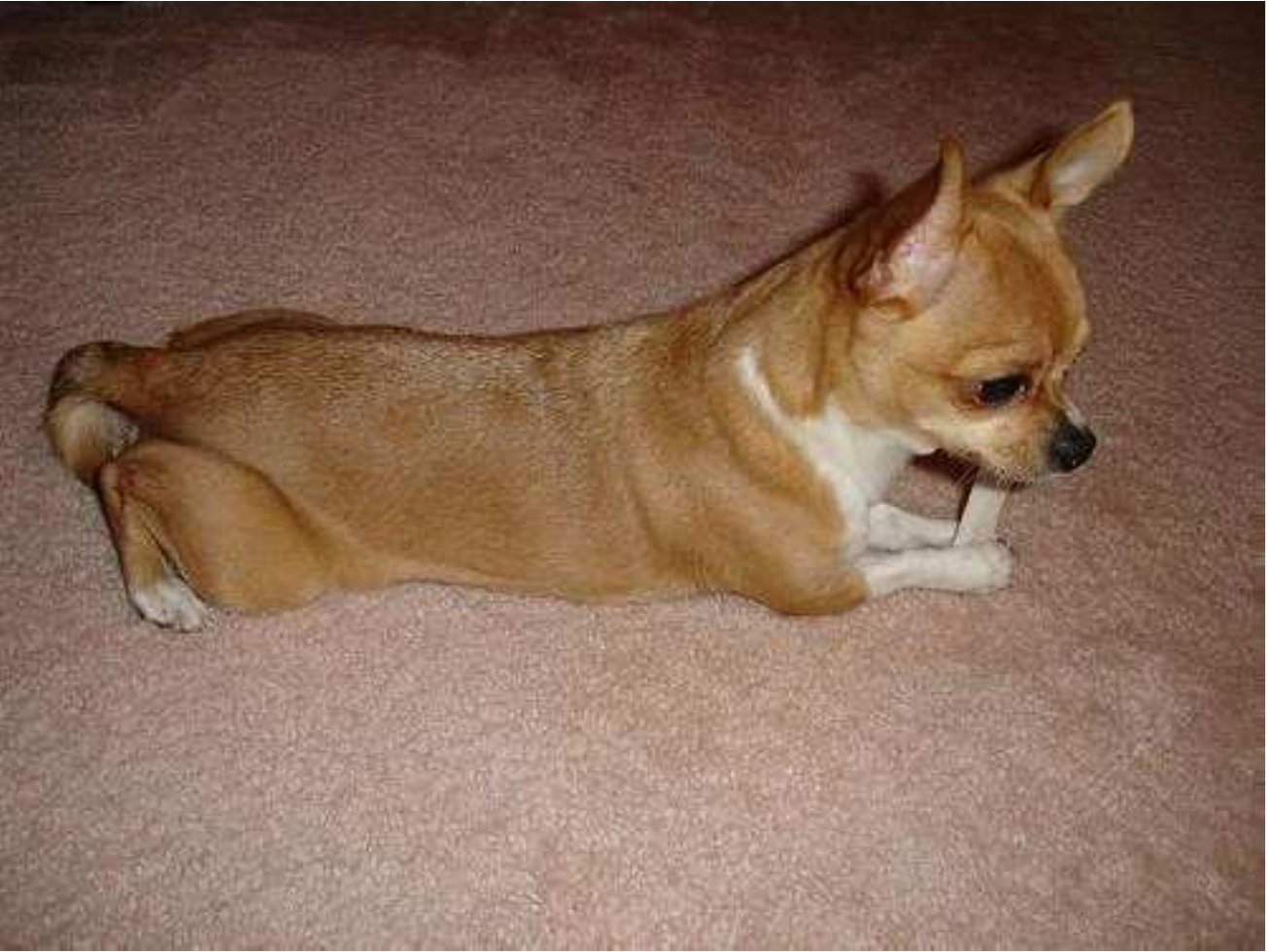}} 
 	
 	\subfloat[]{\label{fig:Sample3}\includegraphics[width=0.42\textwidth]{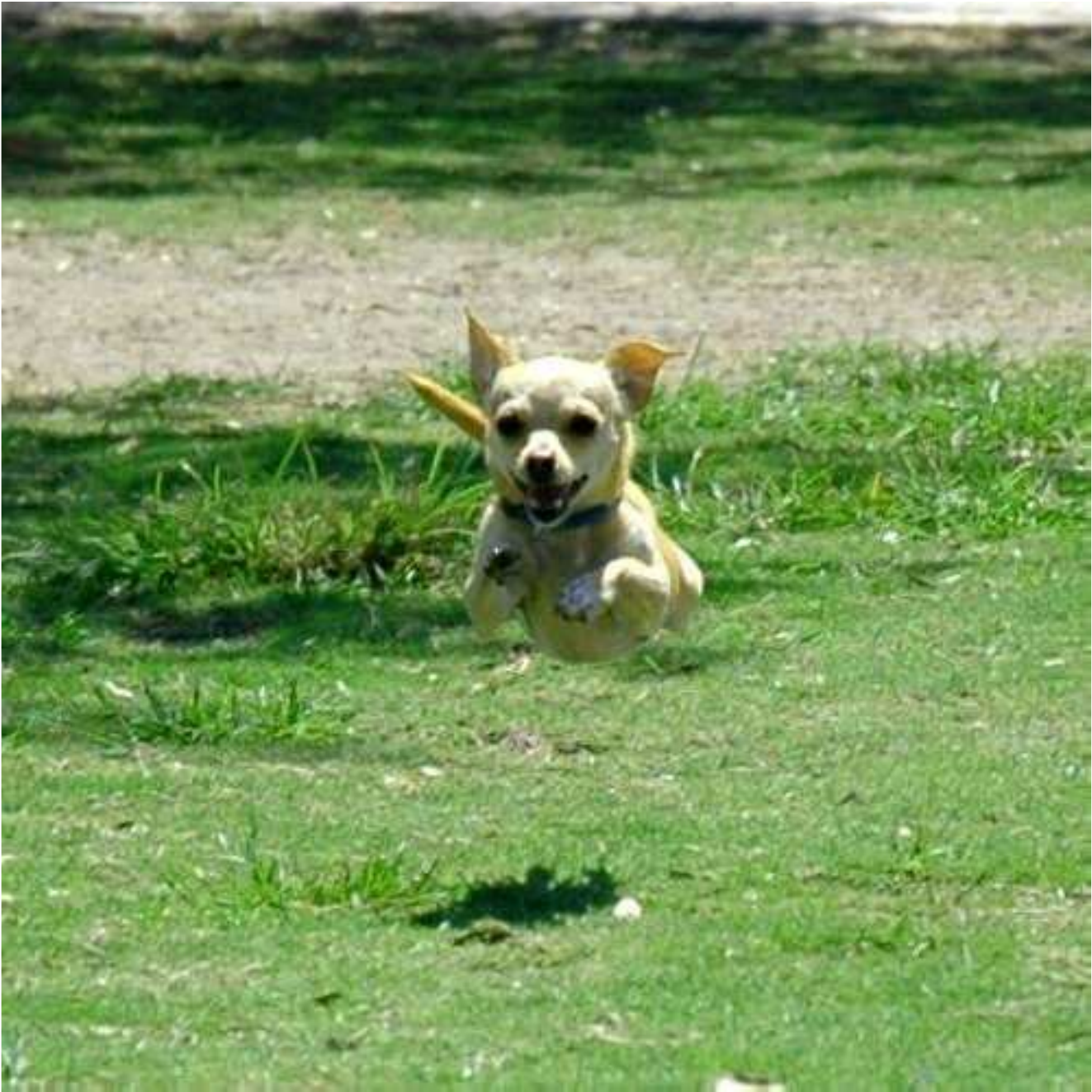}} 
 	\quad
 	\subfloat[]{\label{fig:Sample4}\includegraphics[width=0.42\textwidth]{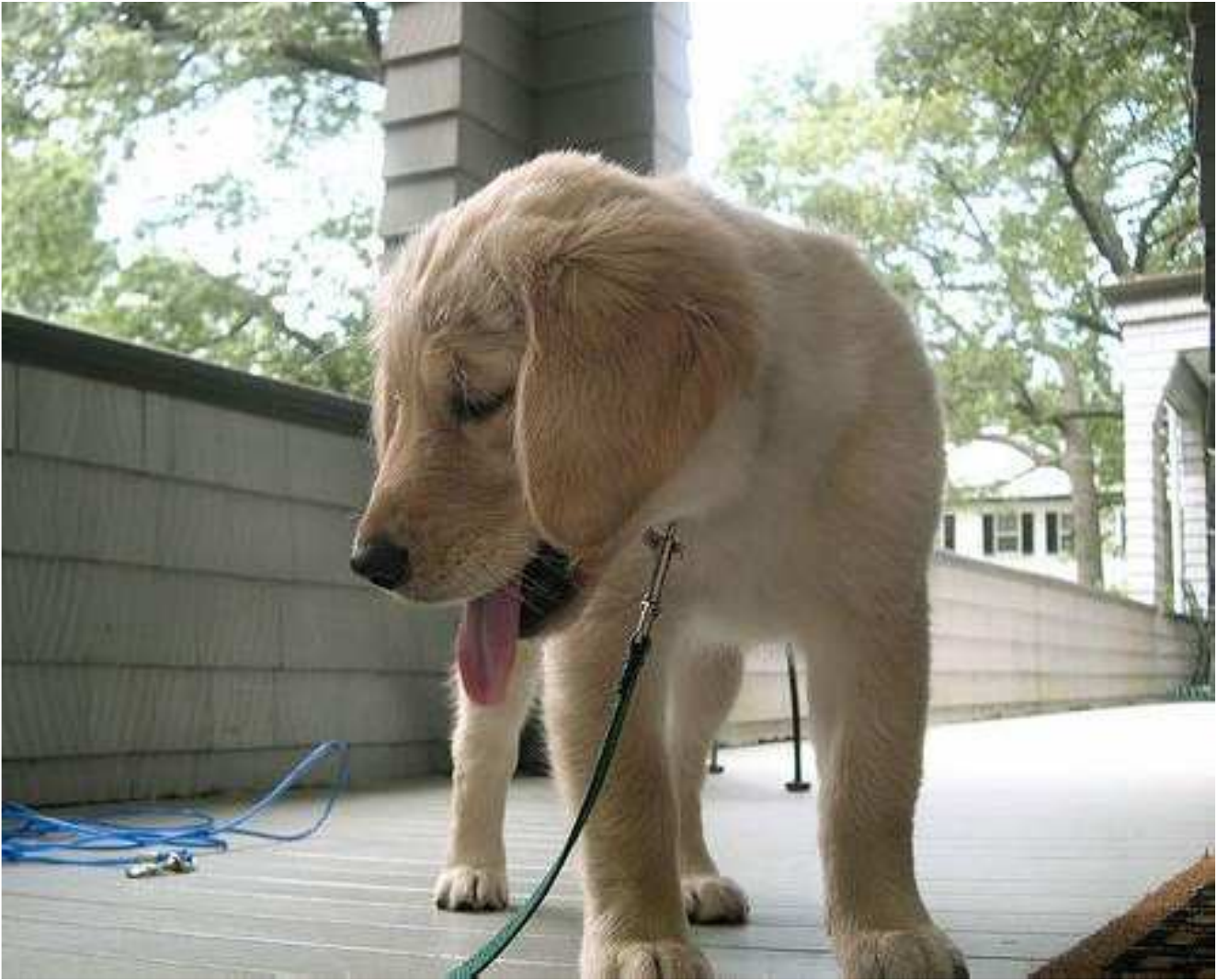}}  
	 \end{minipage}
 \begin{minipage}{.45\linewidth}
 	\subfloat[KPCA projection]{\label{fig:KPCAprojection}\includegraphics[width=1\textwidth]{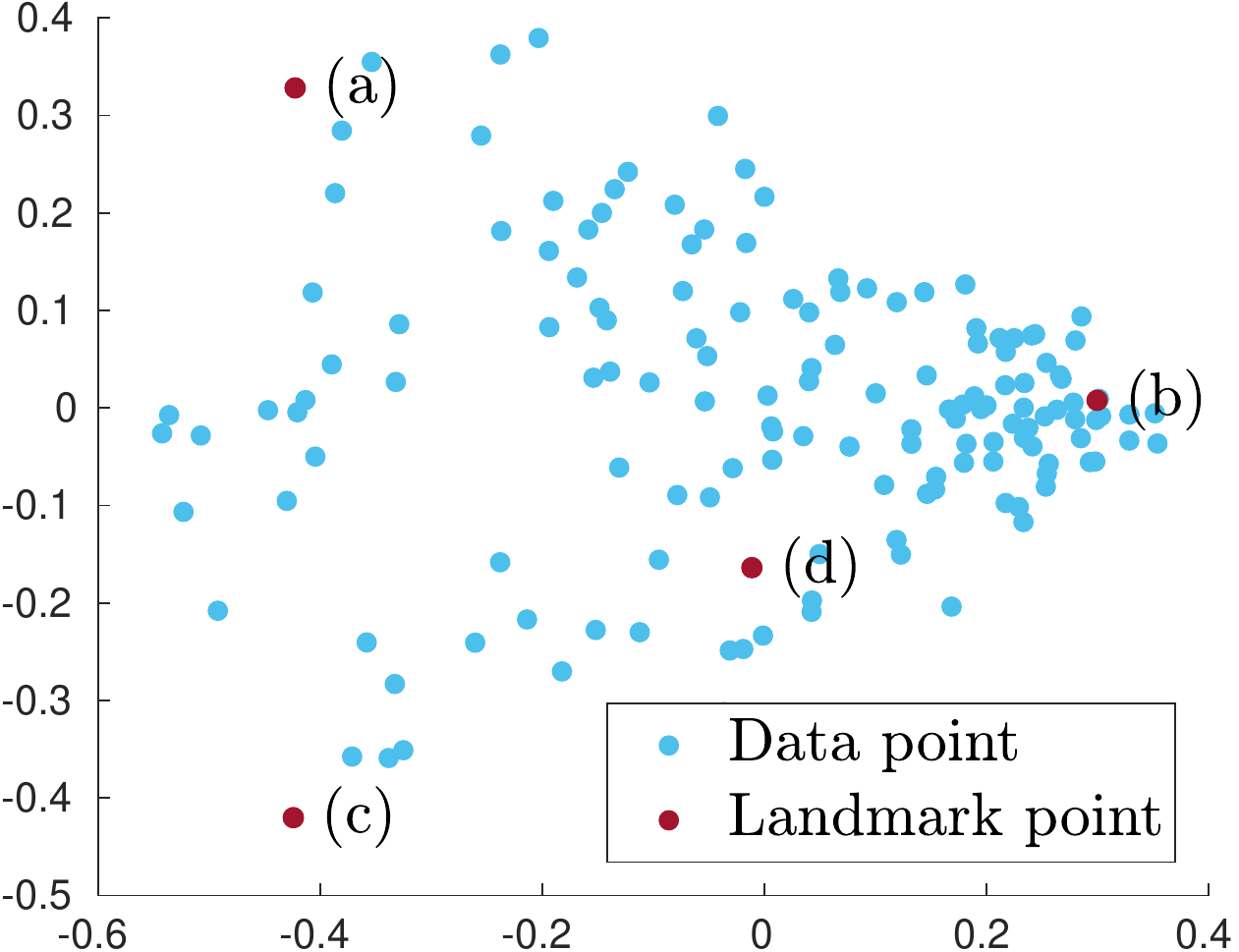}}   
 \end{minipage}  
 	\caption{Illustration of the proposed method with $k=4$ on the \texttt{Stanford Dogs} dataset. The selected landmark points are visualized on the left, the projection on the 2 first principal components of the KPCA on the right.}\label{fig:Dogs}
 \end{figure}
 
\begin{figure}[H]
  	\centering
  	\begin{minipage}{.45\linewidth}
  		\subfloat[]{\label{fig:Sample_Shapes_1}\includegraphics[width=0.22\textwidth]{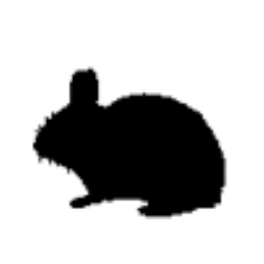}}
  		\quad  
  		\subfloat[]{\label{fig:Sample_Shapes_2}\includegraphics[width=0.22\textwidth]{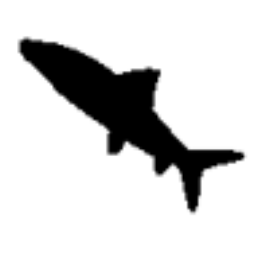}} 
  		\quad	
  		\subfloat[]{\label{fig:Sample_Shapes_3}\includegraphics[width=0.22\textwidth]{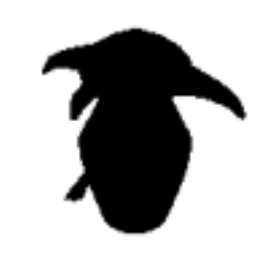}} 
  		
		\subfloat[]{\label{fig:Sample_Shapes_4}\includegraphics[width=0.22\textwidth]{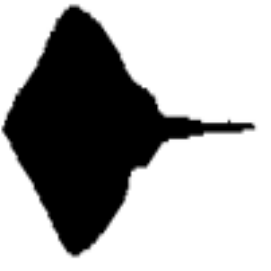}}
		 \quad  
		\subfloat[]{\label{fig:Sample_Shapes_5}\includegraphics[width=0.22\textwidth]{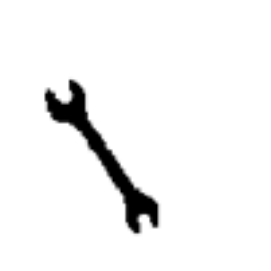}} 
		 \quad	
		\subfloat[]{\label{fig:Sample_Shapes_6}\includegraphics[width=0.22\textwidth]{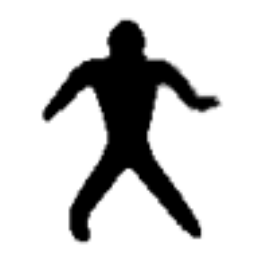}}
		   
		\subfloat[]{\label{fig:Sample_Shapes_7}\includegraphics[width=0.22\textwidth]{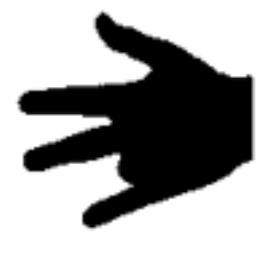}}
		 \quad  
		\subfloat[]{\label{fig:Sample_Shapes_8}\includegraphics[width=0.22\textwidth]{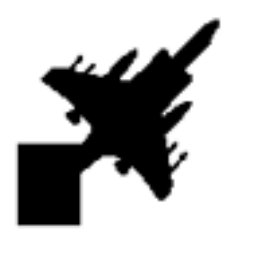}} 
		 \quad	
		\subfloat[]{\label{fig:Sample_Shapes_9}\includegraphics[width=0.22\textwidth]{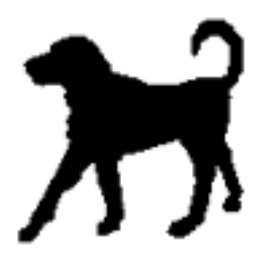}} 
		 
  	\end{minipage}
  	\begin{minipage}{.45\linewidth}
  		\subfloat[KPCA projection]{\label{fig:ProjectionShapes}\includegraphics[width=1\textwidth]{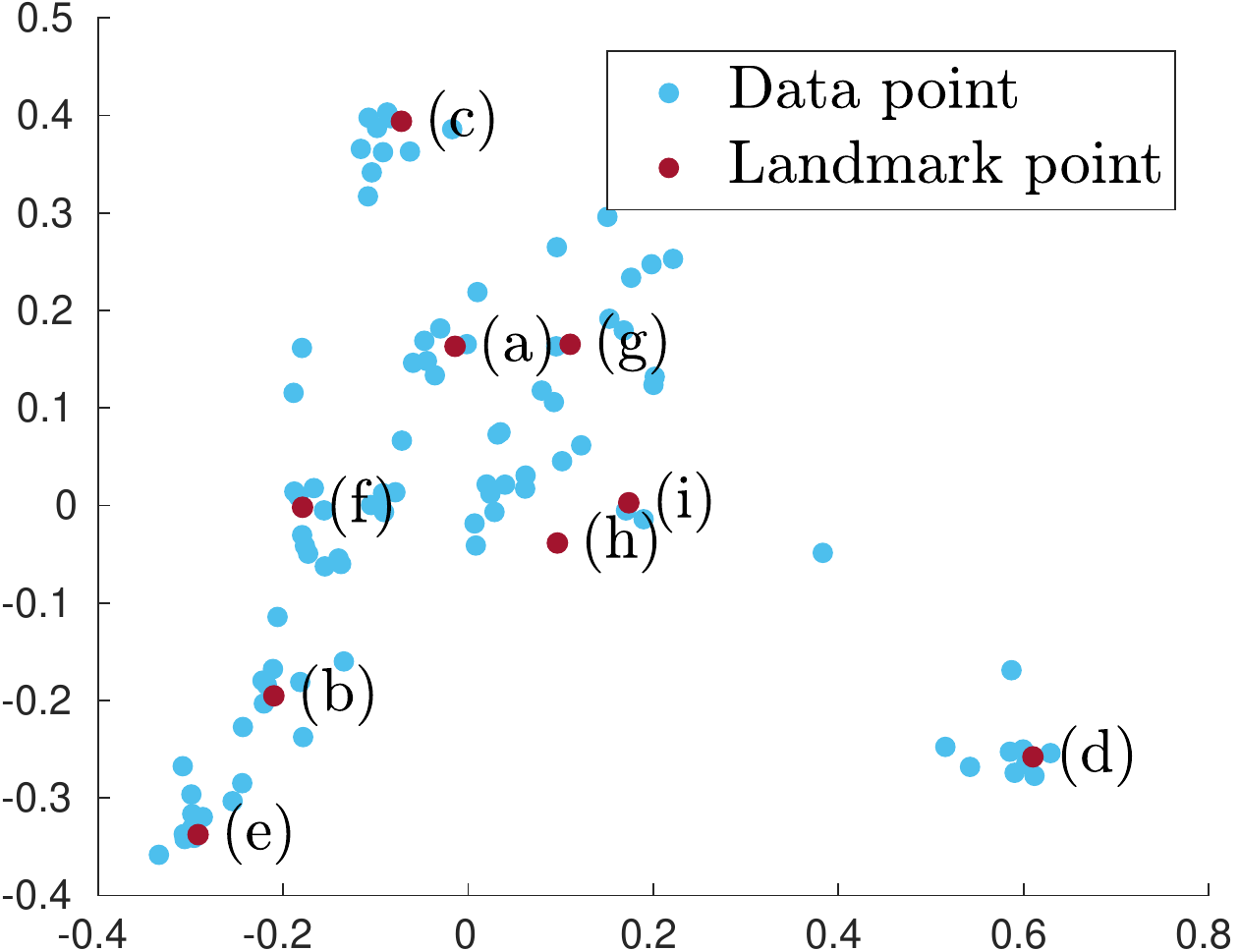}}   
  	\end{minipage}  
  	\caption{Illustration of the proposed method with $k=9$ on the \texttt{Kimia99} dataset. The selected landmark points are visualized on the left, the projection on the 2 first principal components of the KPCA on the right.}\label{fig:Shapes}
  \end{figure}

 We repeat the above procedure on the \texttt{Kimia99} dataset\footnote{\url{https://vision.lems.brown.edu/content/available-software-and-databases\#Datasets-Shape}}. The dataset has 9 classes consisting of 11 images each. It contains shapes silhouettes for the classes: rabbits, quadrupeds, men, airplanes, fish, hands, rays, tools, and a miscellaneous class. The total training set is visible in Figure~\ref{fig:TrainDataKIMIA99} in the Appendix. First, we resize the images to size $100 \times 100$. Afterwards, we apply a Gaussian kernel with bandwidth $\sigma = 100$ after standardizing the data. The results of the proposed method with $k=9$ are visible in Figure~\ref{fig:Shapes}. The method samples landmarks out every class, making it a desirable image summarization. This is supported by the projection of the landmarks on the 2 first principal components of the KPCA, where a landmark points is chosen out of every small cluster.
 
\section{Conclusion}

We discussed a greedy deterministic adaptation of k-DPP. Algorithmically, the method corresponds to the DAS algorithm with a different projector kernel matrix. The proposed method is evaluated by comparing the accuracy of the Nystr\"om approximation on multiple datasets. Experiments show the proposed method is able to give a more diverse subset, along with better performance for the relative max norm. When there is a fast decay of the eigenvalues, the deterministic method is more accurate than randomized counterparts. To conclude, we demonstrate the usefulness of the model on an image search task.

\subsection*{Acknowledgements}
\footnotesize{
EU: The research leading to these results has received funding from
the European Research Council under the European Union's Horizon
    2020 research and innovation program / ERC Advanced Grant E-DUALITY
    (787960). This paper reflects only the authors' views and the Union
    is not liable for any use that may be made of the contained information.
Research Council KUL: Optimization frameworks for deep kernel machines C14/18/068
Flemish Government:
FWO: projects: GOA4917N (Deep Restricted Kernel Machines:
        Methods and Foundations), PhD/Postdoc grant
Impulsfonds AI: VR 2019 2203 DOC.0318/1QUATER Kenniscentrum Data
        en Maatschappij
Ford KU Leuven Research Alliance Project KUL0076 (Stability analysis
    and performance improvement of deep reinforcement learning algorithms). 
}
%
%
%
%

\clearpage
\bibliographystyle{unsrt}
\bibliography{References}

\appendix

\section{Additional algorithms}\label{app:algo}

\begin{algorithm}[h]
\begin{algorithmic}
\Statex {\bf input}: Matrix $K\succ 0$, sample size $k$ and  $\gamma > 0$.
\Statex {\bf initialization}: $\mathcal{C}_0 = \emptyset$ and $m=1$.
\Statex  $P \leftarrow  K(K+n\gamma \mathbb{I})^{-1}$
\Statex {\bf while}: $m\leq k$ \, {\bf do}
\Statex \quad  $s_{m} \in \arg\max \diag\left(P-P_{\mathcal{C}_m}P_{\mathcal{C}_m\mathcal{C}_m}^{-1}P^\top_{\mathcal{C}_m}\right)$. 
\Statex \quad  $\mathcal{C}_{m} \leftarrow \mathcal{C}_{m-1}\cup\{s_m\}$ and $m\leftarrow m+1$.
\Statex {\bf end while}
\Statex {\bf return} $\mathcal{C}_m$.
\end{algorithmic} 
\caption{DAS algorithm~\cite{RegularizedChristoffel}.\label{DAS}}
\end{algorithm}

\section{Additional figures}\label{app:AddFig}

\begin{figure}[h]
	\centering
	\includegraphics[width=1\textwidth]{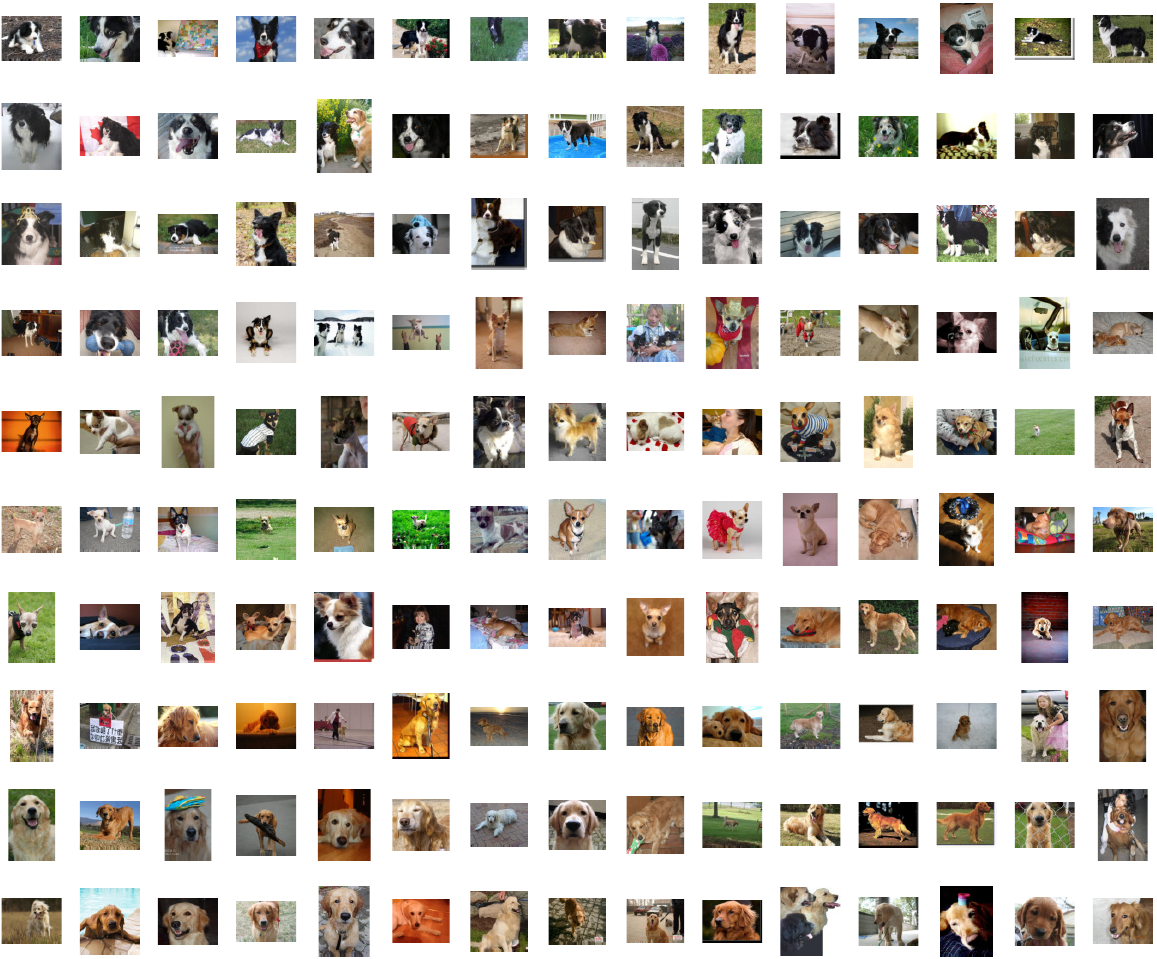}
	\caption{The training data of the \texttt{Stanford Dogs} dataset.}
	\label{fig:TrainData}
\end{figure}

\begin{figure}[h]
	\centering
	\includegraphics[width=1\textwidth]{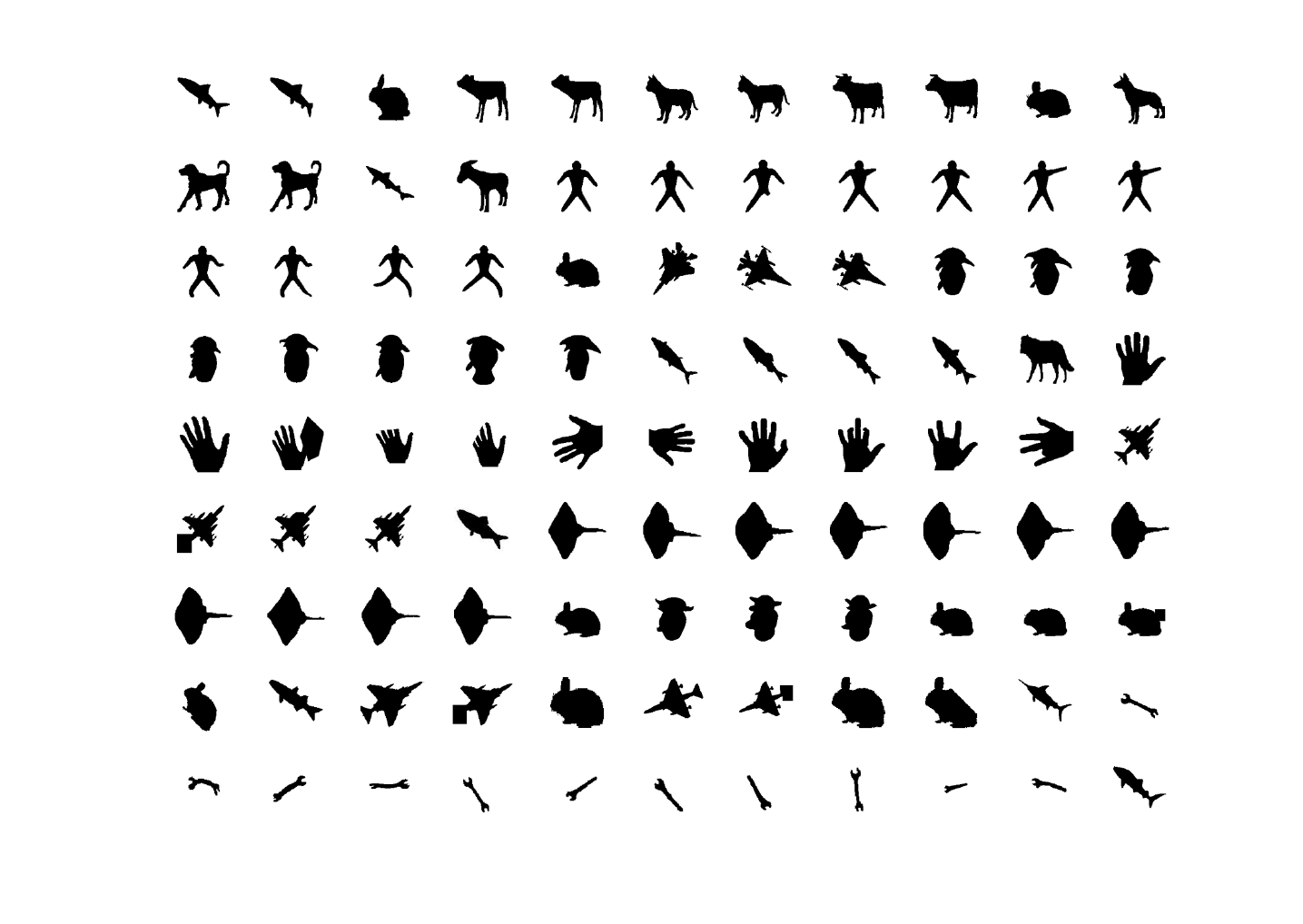}
	\caption{The training data of the \texttt{Kimia99} dataset.}
	\label{fig:TrainDataKIMIA99}
\end{figure}

\begin{figure}[h]
	\centering
\subfloat[\texttt{Stock}]{\label{fig:Stock_eig}\includegraphics[width=0.24\textwidth]{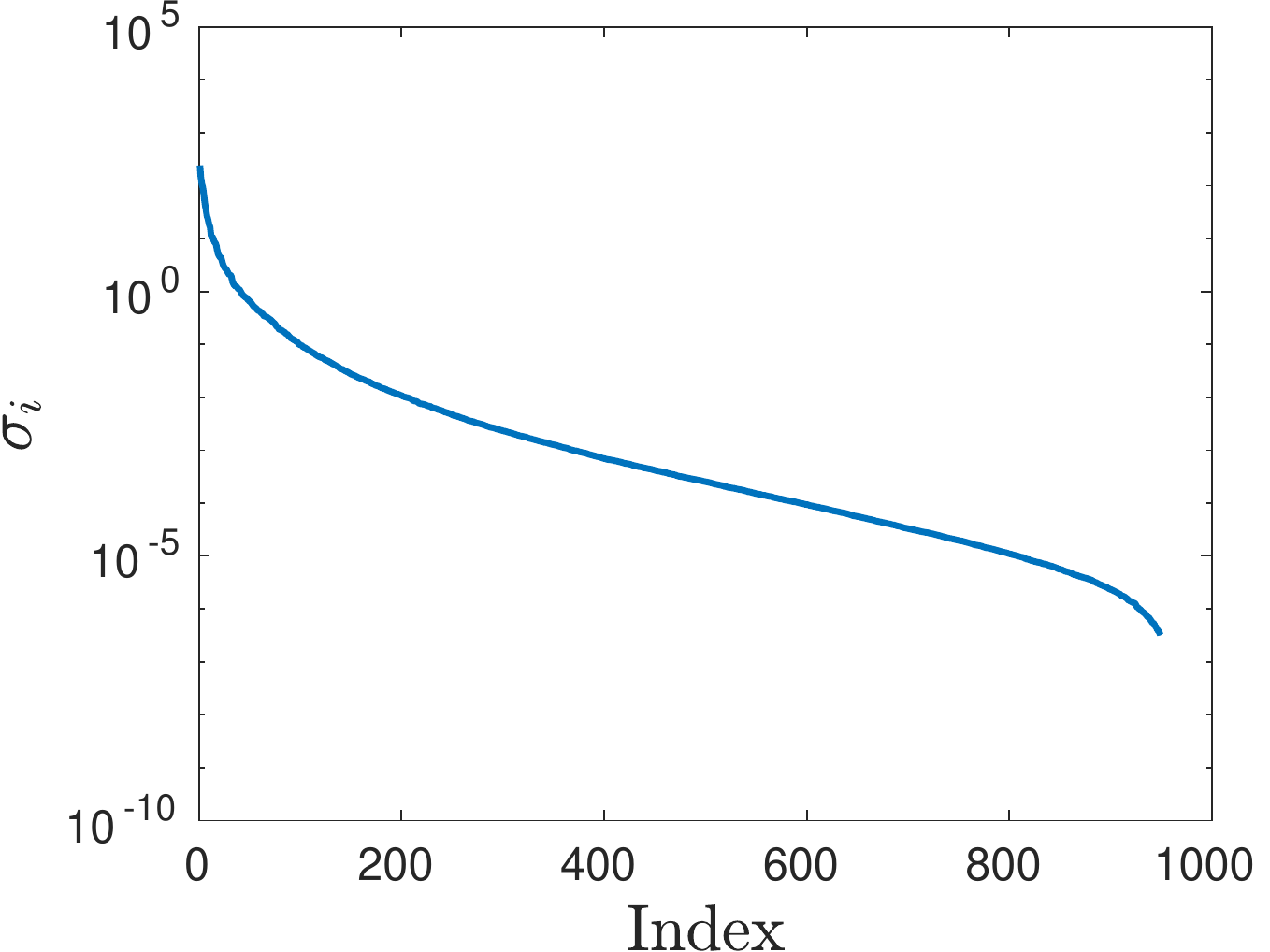}}  
\subfloat[\texttt{Housing}]{\label{fig:Housing_eig}\includegraphics[width=0.24\textwidth]{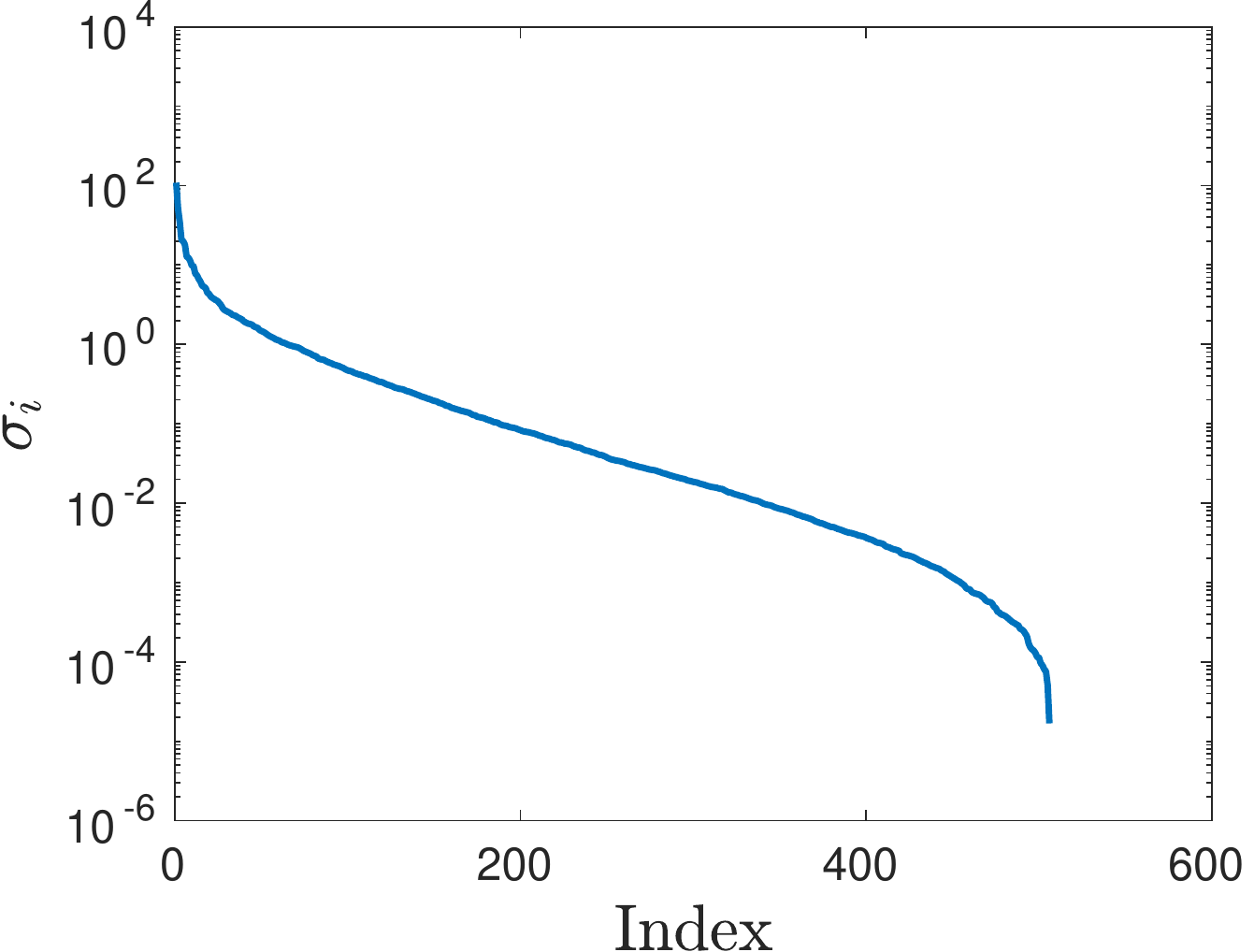}}
\subfloat[\texttt{Abalone}]{\label{fig:Abalone_eig}\includegraphics[width=0.24\textwidth]{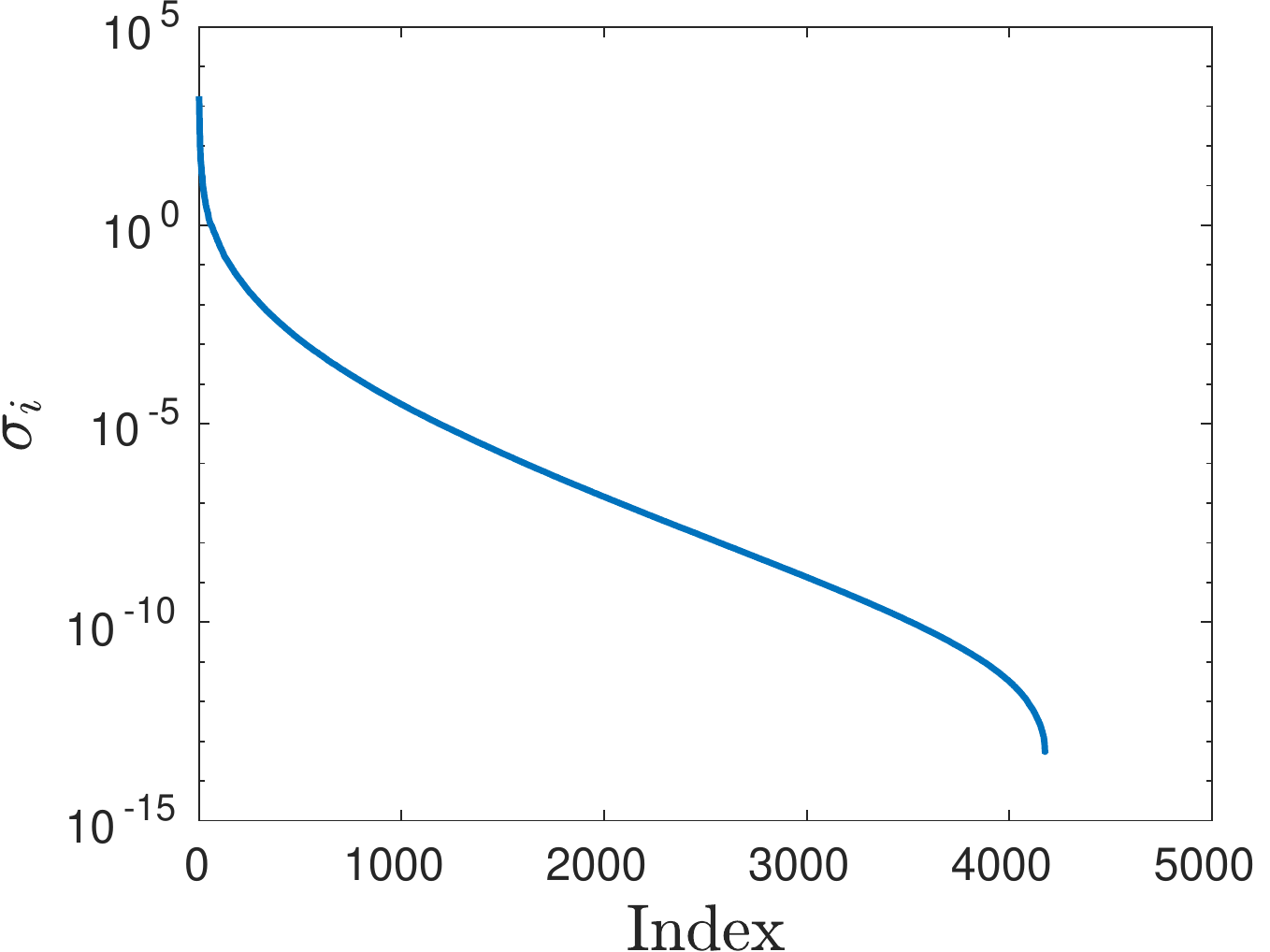}}
\subfloat[\texttt{Bank8FM}]{\label{fig:Bank8FM_eig}\includegraphics[width=0.24\textwidth]{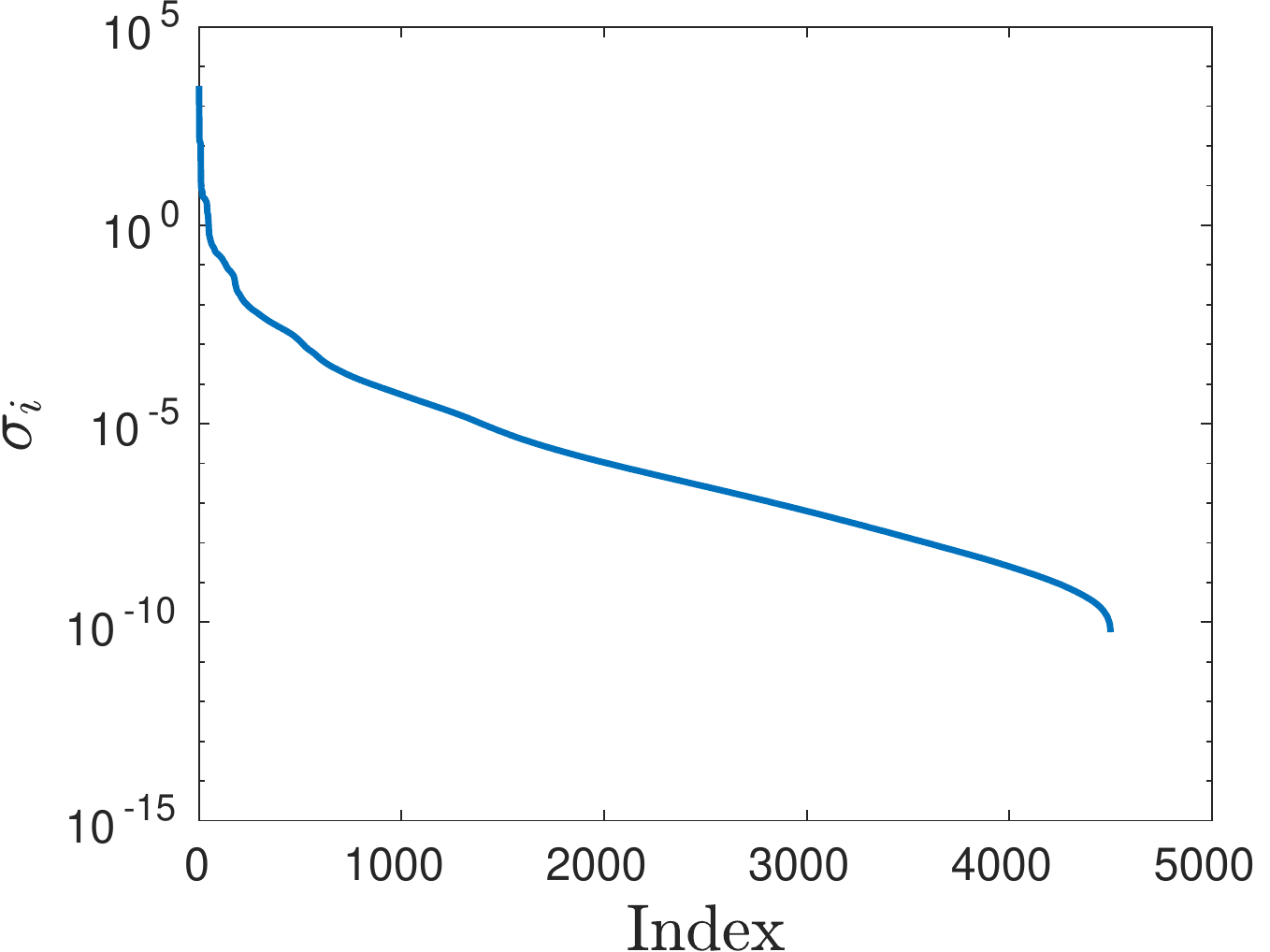}}
	\caption{Singular value spectrum of the datasets on a logarithmic scale. For a given index, the value of the eigenvalues for the \texttt{Stock} and \texttt{Housing} dataset are smaller than \texttt{Abalone} and \texttt{Bank8FM}.}\label{fig:Eig}
\end{figure}

  \begin{figure}[h]
 	\centering
 	\subfloat[\texttt{Stock}]{\label{fig:stock_Det}\includegraphics[width=0.45\textwidth]{Figures/stock_Det.eps}}   
 	\quad
 	\subfloat[\texttt{Housing}]{\label{fig:housing_Det}\includegraphics[width=0.45\textwidth]{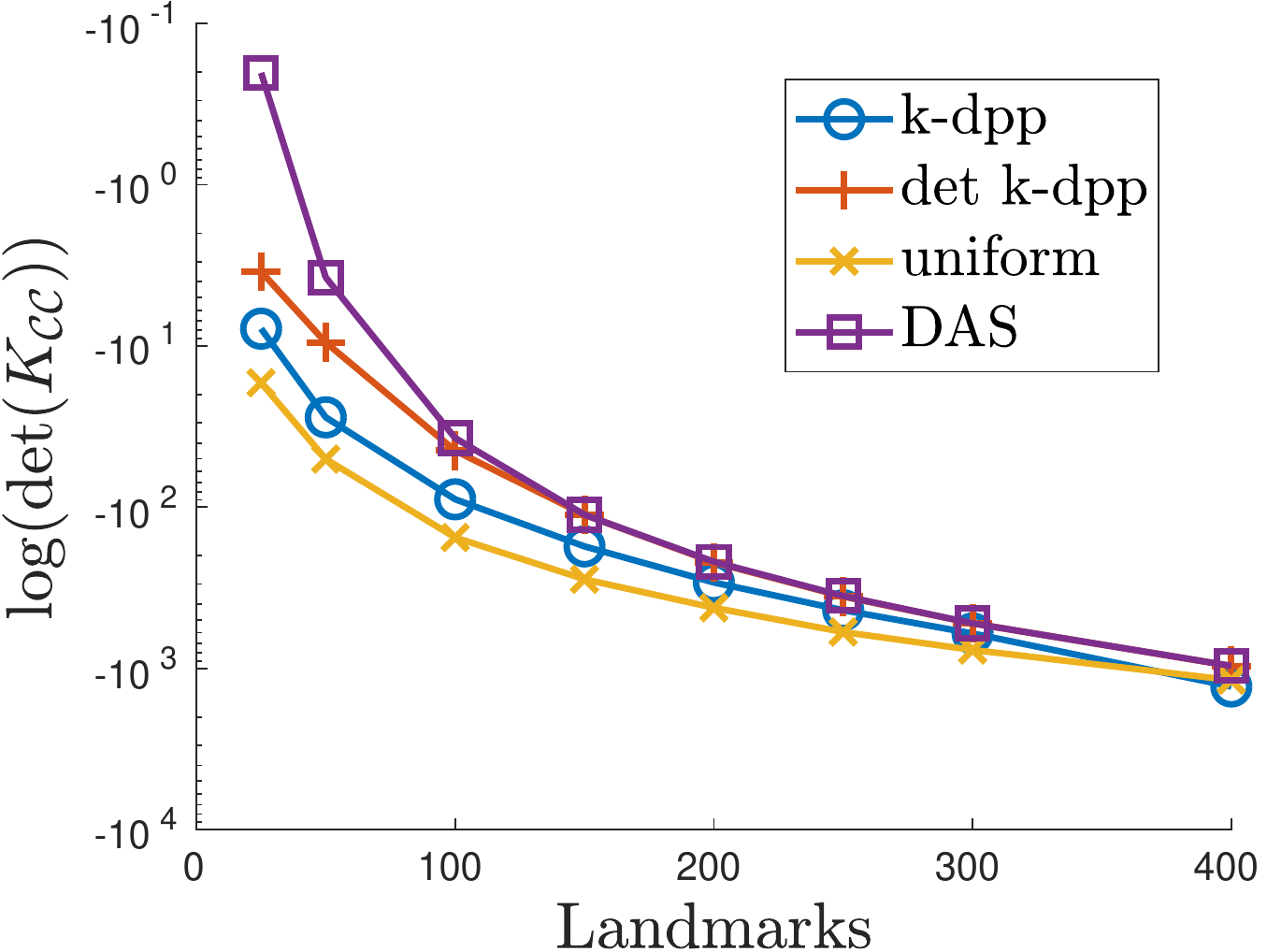}}   
 	\quad
 	\subfloat[\texttt{Abalone}]{\label{fig:abalone_Det}\includegraphics[width=0.45\textwidth]{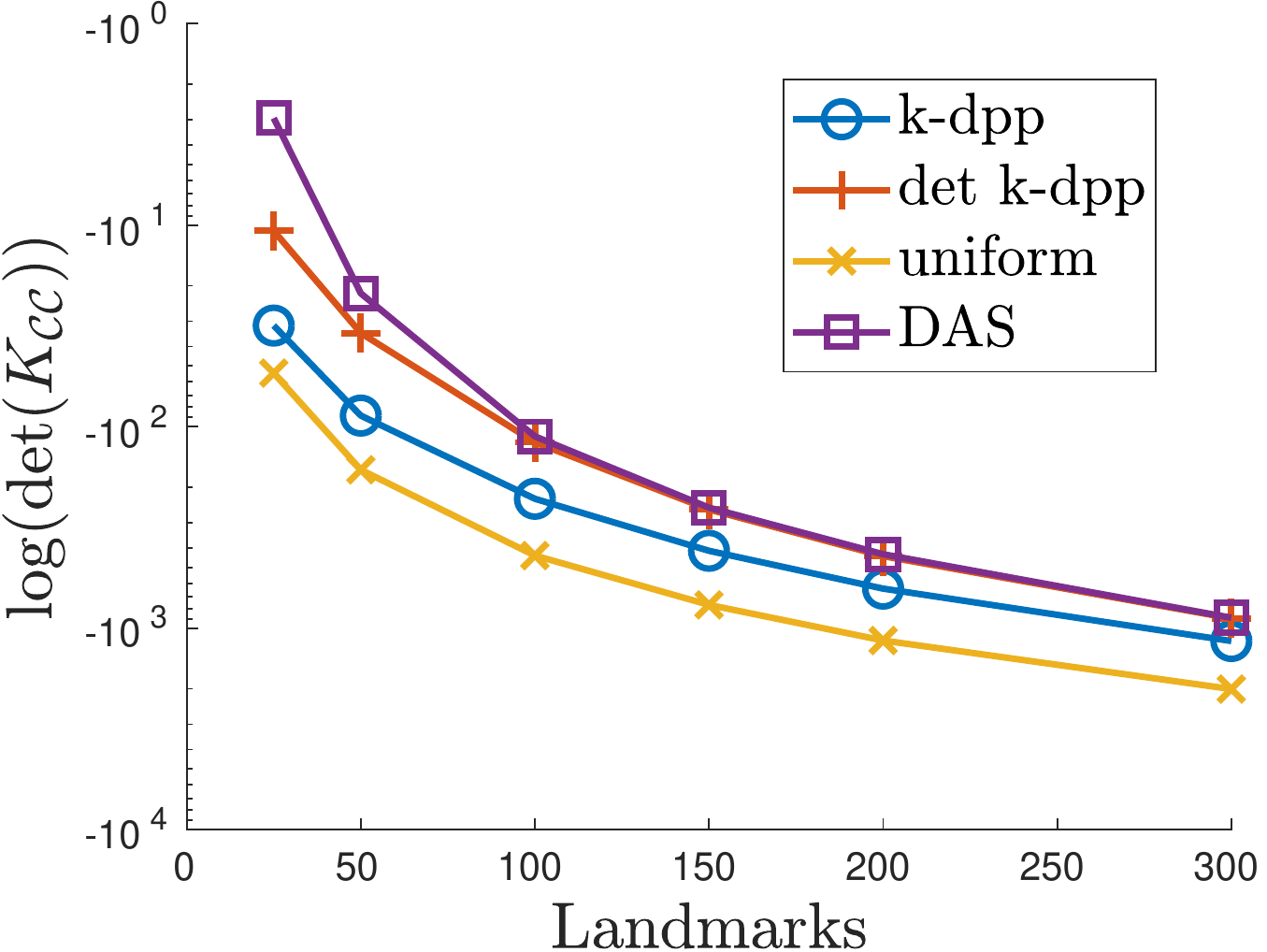}}   
 	\quad
 	\subfloat[\texttt{Bank8FM}]{\label{fig:bank8FM_Det}\includegraphics[width=0.45\textwidth]{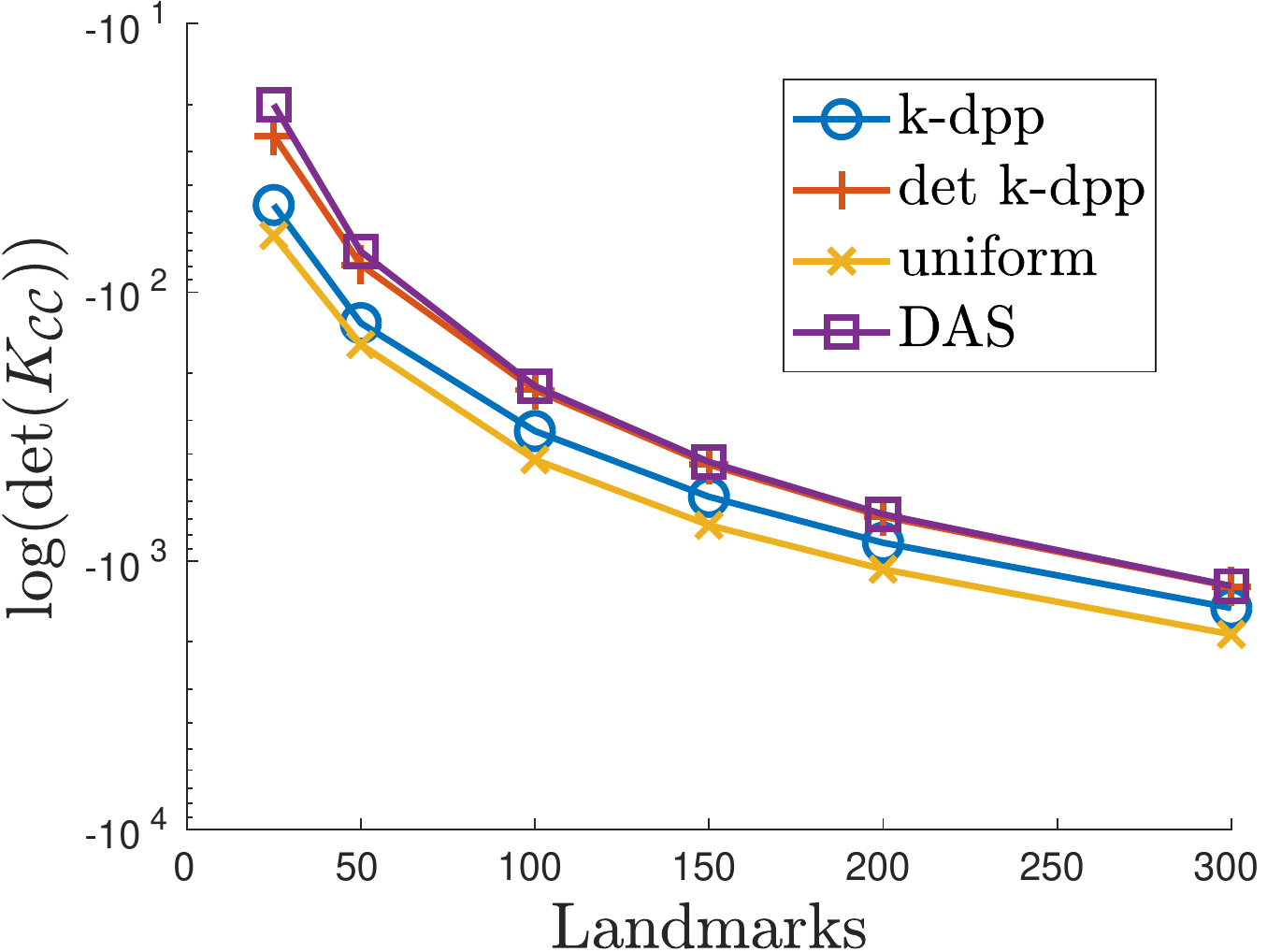}}  
 	\caption{$\mathrm{log}(\mathrm{det}(K_{\mathcal{C}\mathcal{C}}))$ in function of the number of landmarks. The error is plotted on a logarithmic scale, averaged over 10 trials. The larger the $\mathrm{log}(\mathrm{det}(K_{\mathcal{C}\mathcal{C}}))$, the more diverse the subset.}\label{fig:Det}
 \end{figure}
 
 \begin{figure}[h]
 	\centering
 	\subfloat[\texttt{Stock}]{\label{fig:Stock_OP}\includegraphics[width=0.45\textwidth]{Figures/stock_Op.eps}}   
 	\quad
 	\subfloat[\texttt{Housing}]{\label{fig:Housing_Op}\includegraphics[width=0.45\textwidth]{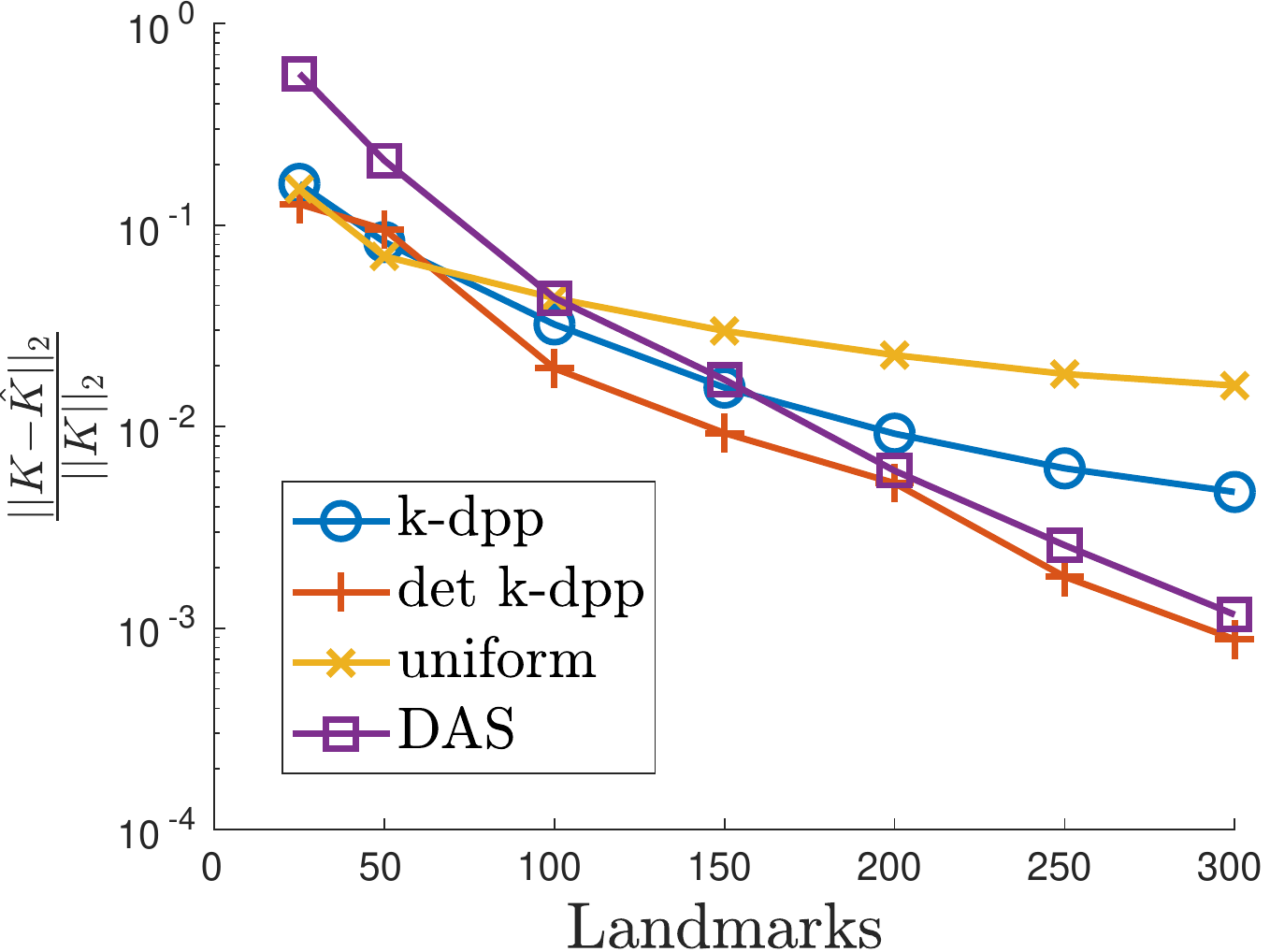}}   
 	\quad
 	\subfloat[\texttt{Abalone}]{\label{fig:Abalone_Op}\includegraphics[width=0.45\textwidth]{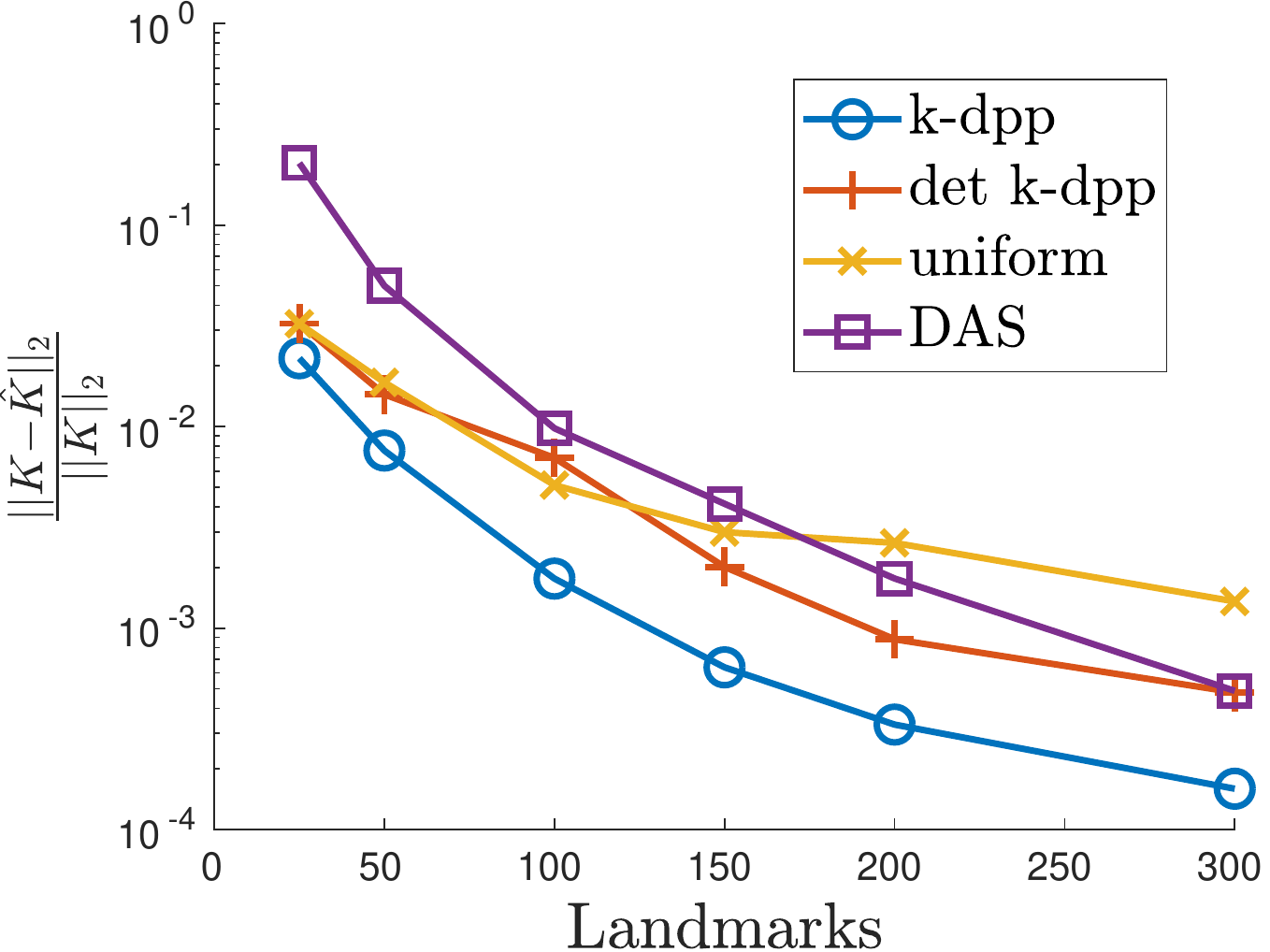}}   
 	\quad
 	\subfloat[\texttt{Bank8FM}]{\label{fig:Bank8FM_Op}\includegraphics[width=0.45\textwidth]{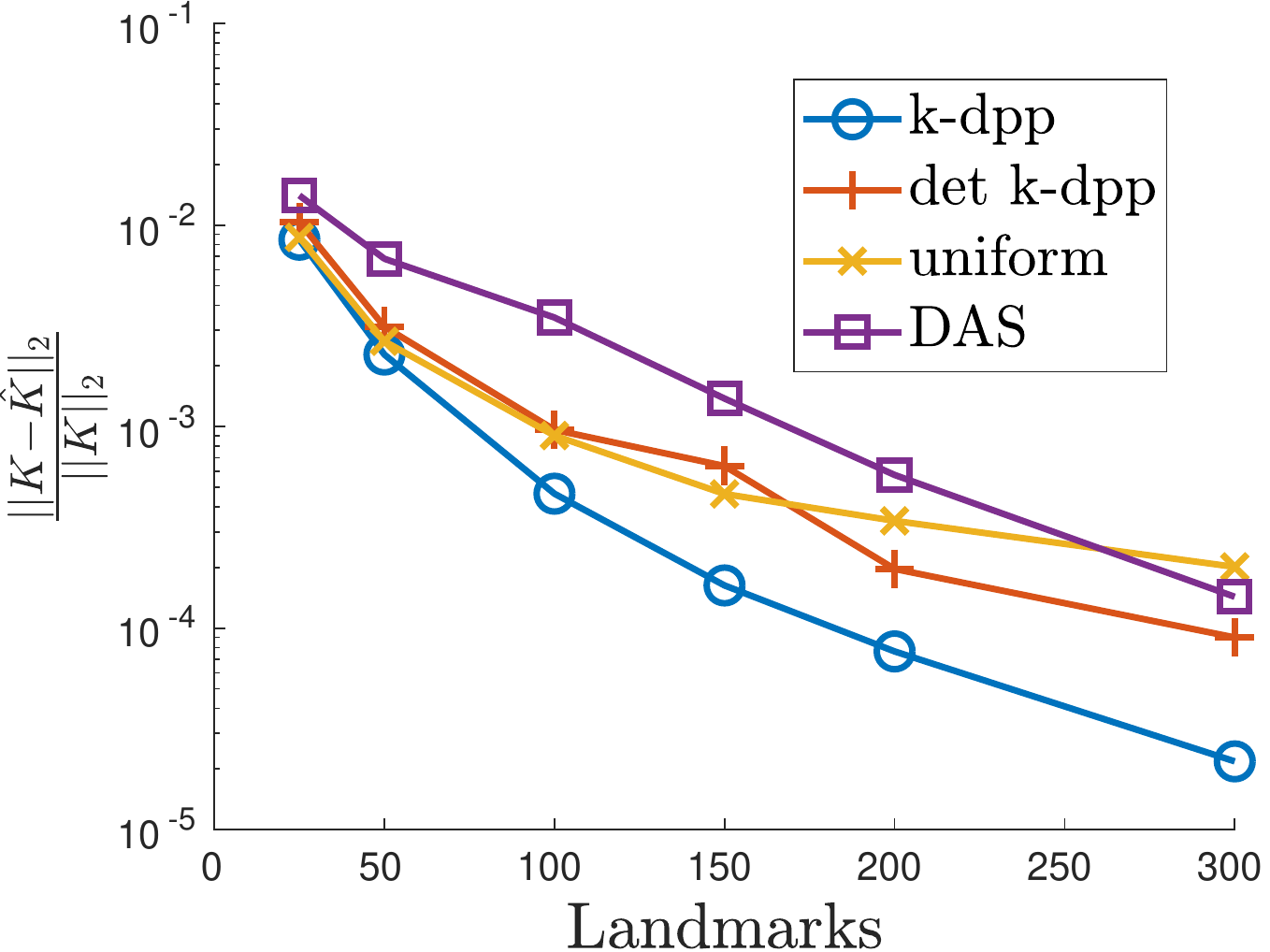}}    
 	\caption{Relative operator norm of the Nystr\"om approximation error as a function of the number of landmarks. The error is plotted on a logarithmic scale, averaged over 10 trials.}\label{fig:OP}
 \end{figure}
 
 \begin{figure}[h]
 	\centering
 	\subfloat[\texttt{Stock}]{\label{fig:stock_Max}\includegraphics[width=0.45\textwidth]{Figures/stock_Max.eps}}   
 	\quad
 	\subfloat[\texttt{Housing}]{\label{fig:housing_Max}\includegraphics[width=0.45\textwidth]{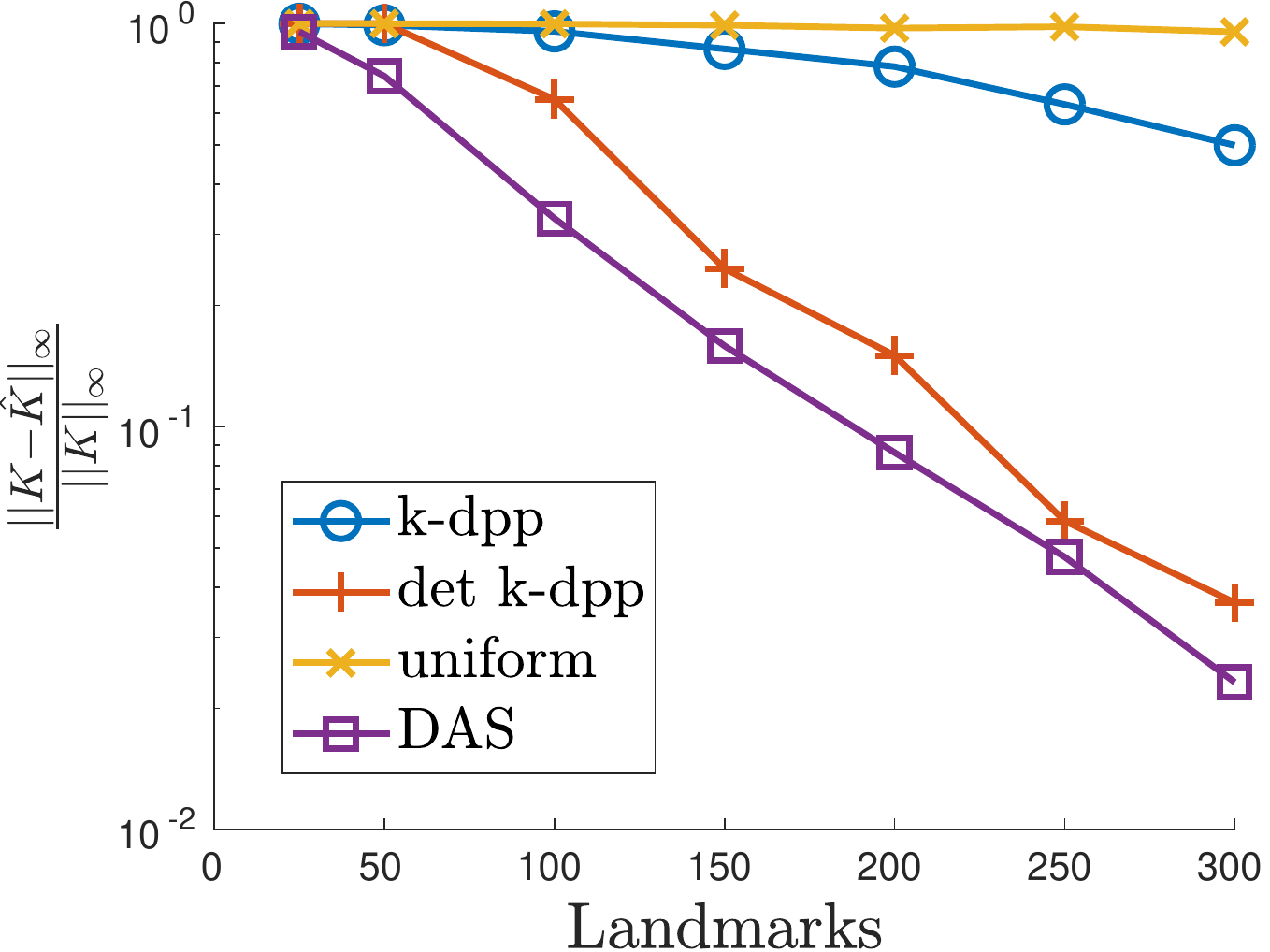}}   
 	\quad
 	\subfloat[\texttt{Abalone}]{\label{fig:abalone_Max}\includegraphics[width=0.45\textwidth]{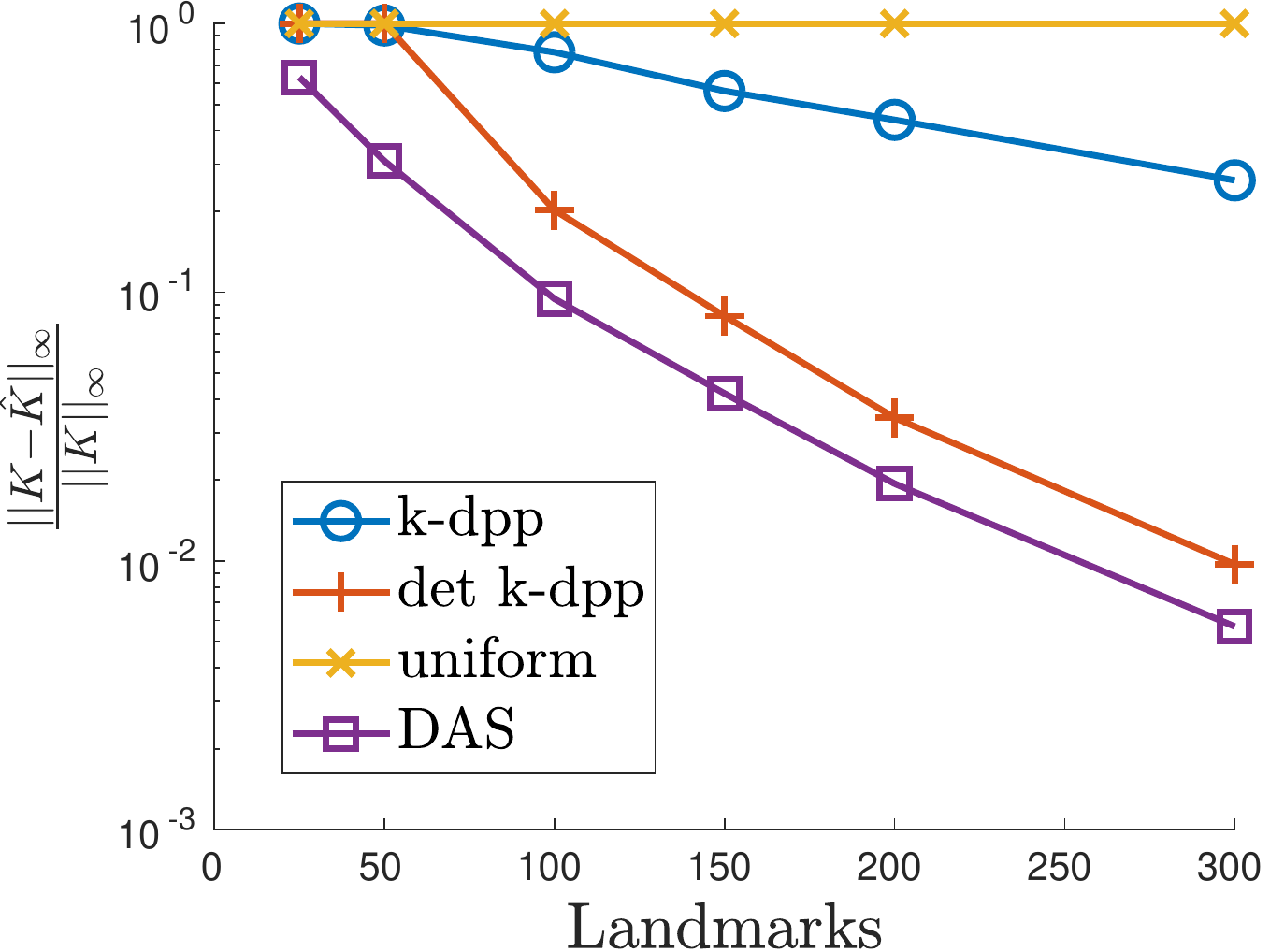}}   
 	\quad
 	\subfloat[\texttt{Bank8FM}]{\label{fig:bank8FM_Max}\includegraphics[width=0.45\textwidth]{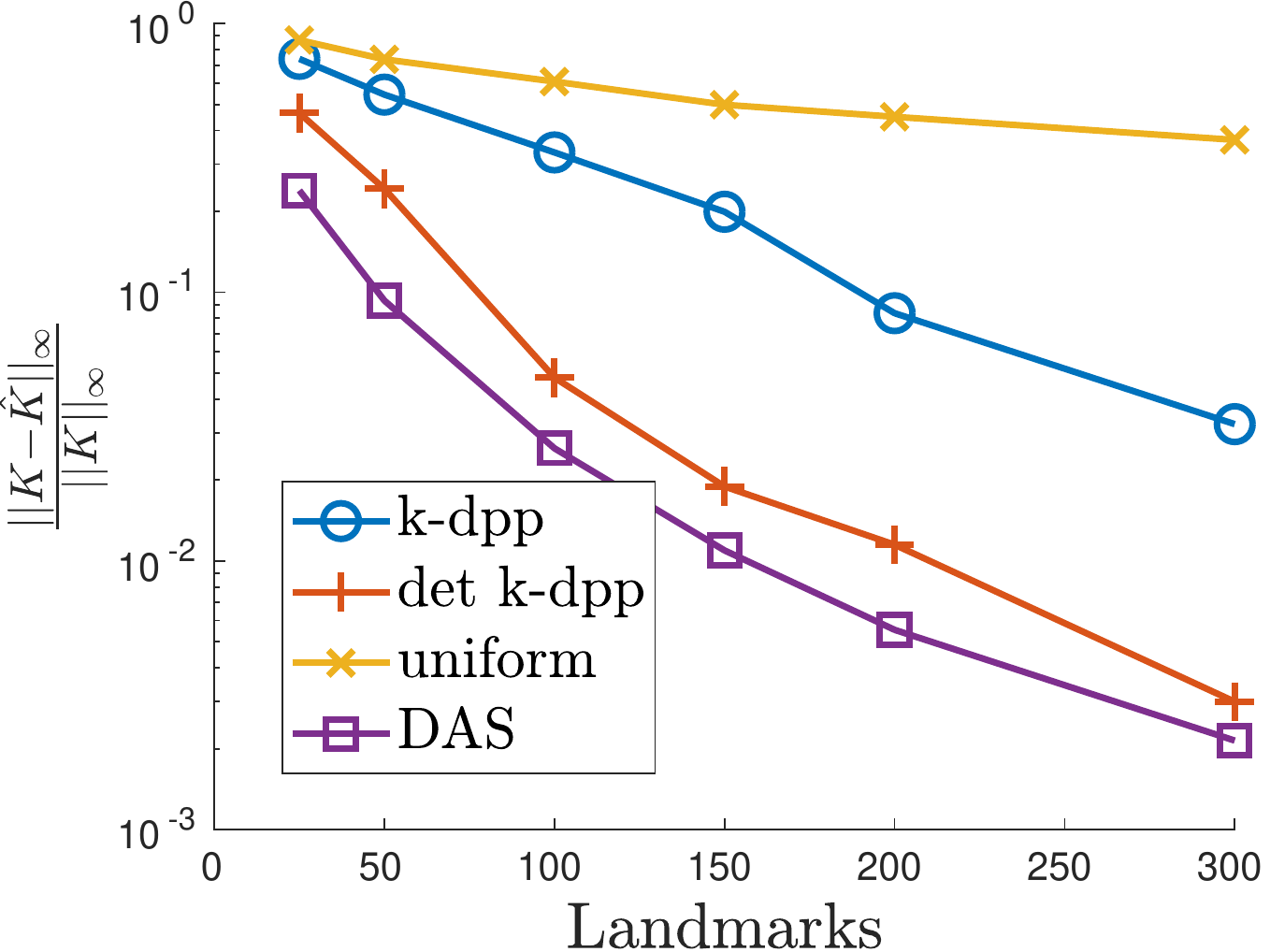}}  
 	\caption{Relative max norm of the approximation as a function of the number of landmarks. The error is plotted on a logarithmic scale, averaged over 10 trials.}\label{fig:Max}
 \end{figure}
 
 \begin{figure}[h]
 	\centering
 	\subfloat[\texttt{Stock}]{\label{fig:stock_Time}\includegraphics[width=0.24\textwidth]{Figures/stock_Time.eps}}   
 	\subfloat[\texttt{Housing}]{\label{fig:housing_Time}\includegraphics[width=0.24\textwidth]{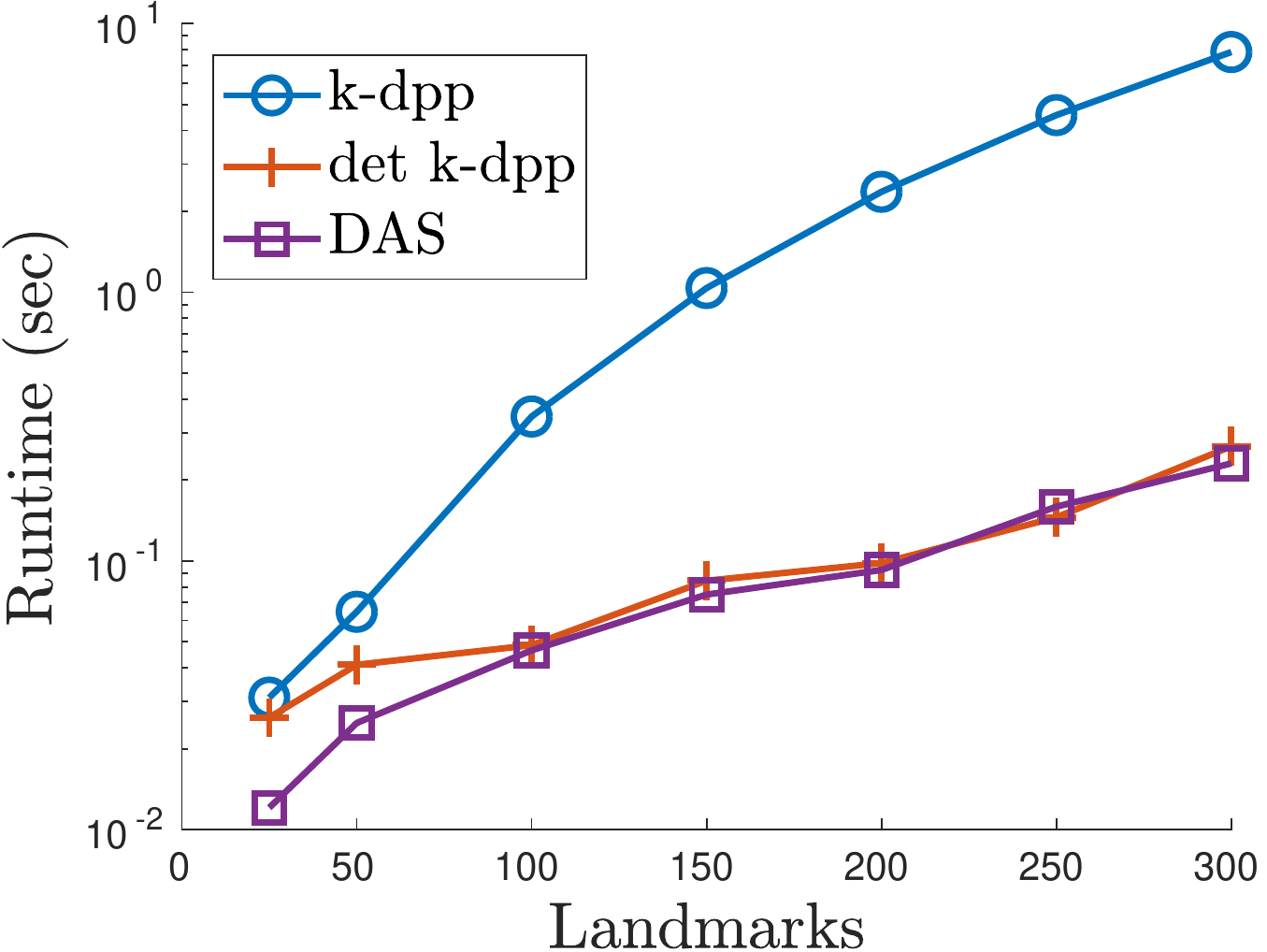}}   
 	\subfloat[\texttt{Abalone}]{\label{fig:abalone_Time}\includegraphics[width=0.24\textwidth]{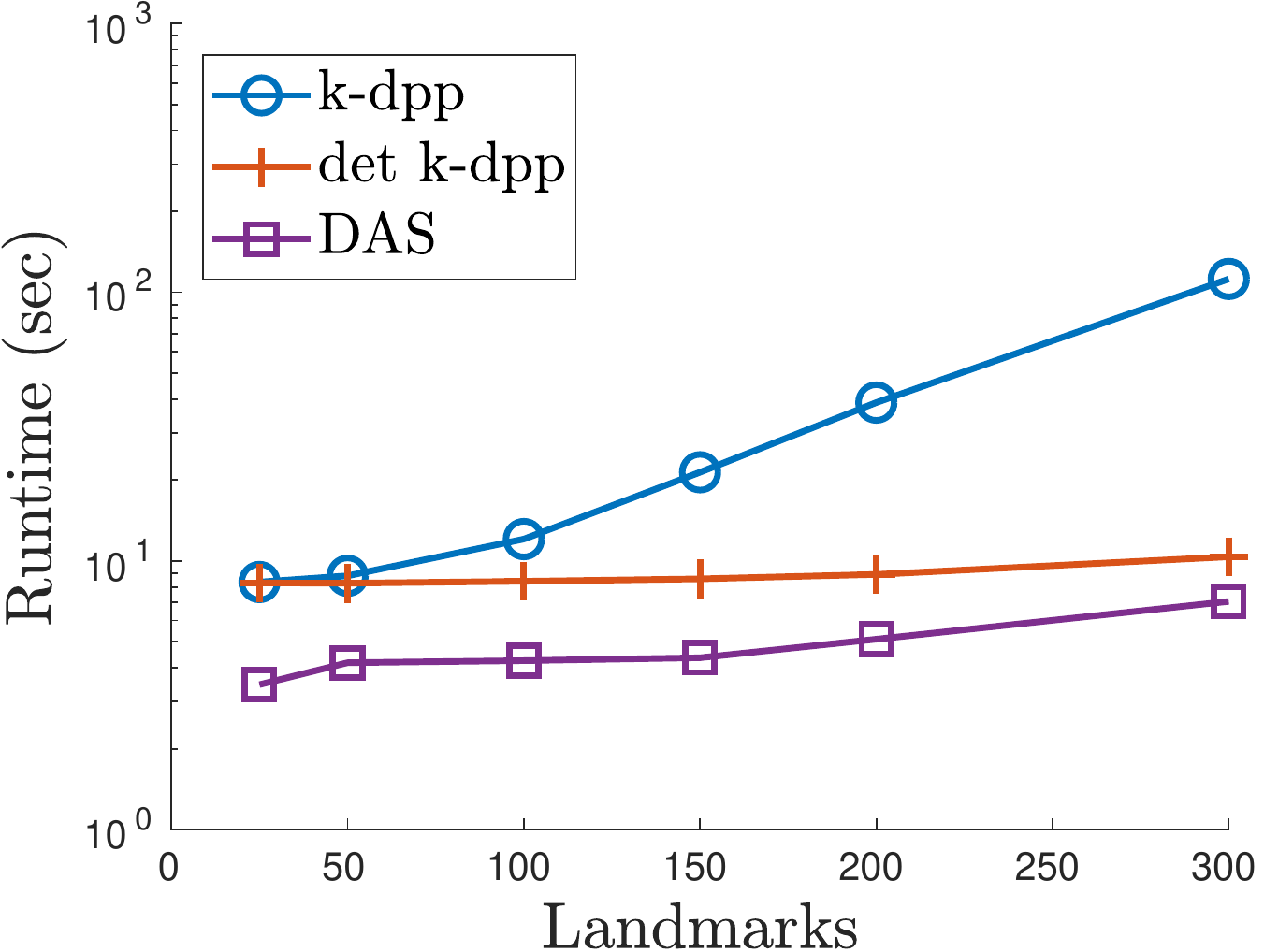}}   
 	\subfloat[\texttt{Bank8FM}]{\label{fig:bank8FM_Time}\includegraphics[width=0.24\textwidth]{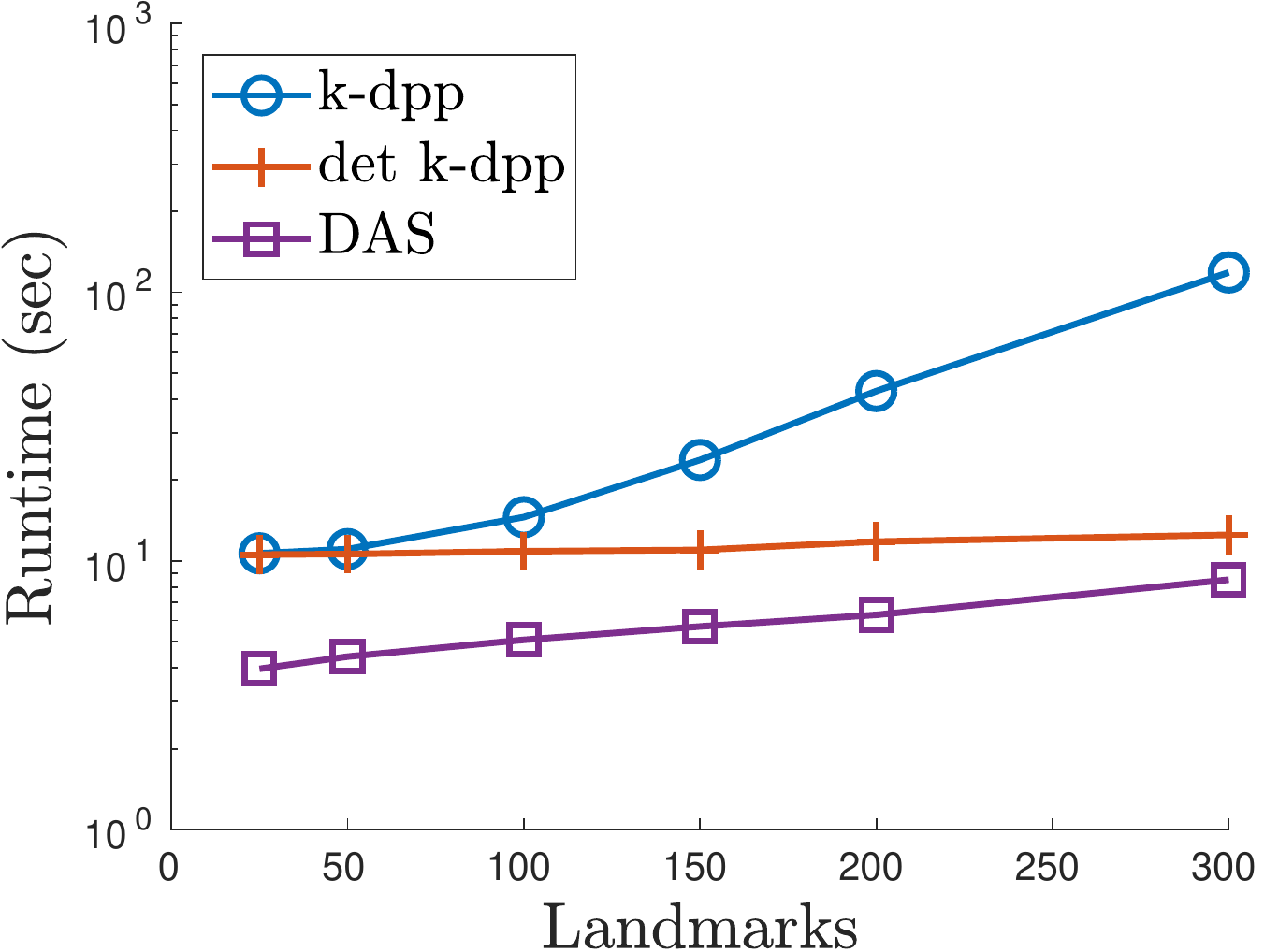}}  
 	\caption{Timings for the computations of Figure~\ref{fig:OP} as a function of the number of landmarks. The timings are plotted on a logarithmic scale, averaged over 10 trials.}\label{fig:Time}
 \end{figure}

\end{document}